\documentclass[sigconf,nonacm]{acmart}

\usepackage{balance} 
\usepackage{algorithm}
\usepackage{algpseudocode}
\usepackage{multirow}

\fancypagestyle{firstpagestyle}{%
  \fancyhf{}%
  \fancyhead[L]{\small Preprint}%
  \fancyfoot[C]{\small\thepage}%
}
\fancypagestyle{standardpagestyle}{%
  \fancyhf{}%
  \fancyhead[L]{\small Preprint}%
  \fancyfoot[C]{\small\thepage}%
}

\title[LEMUR]{LEMUR: Learning to Align with 
\\
Multi-Objective Reinforcement Learning from Preference Feedback}

\author{\texorpdfstring{Manith Adikari\textsuperscript{1,4,*},
        Bei Peng\textsuperscript{2},
        Samuele Vinanzi\textsuperscript{3},
        Angelo Cangelosi\textsuperscript{1,4}}%
        {Manith Adikari, Bei Peng, Samuele Vinanzi, Angelo Cangelosi}}
\affiliation{%
  \institution{\textsuperscript{1}Department of Computer Science, University of Manchester}
  \institution{\textsuperscript{2}School of Computer Science, University of Sheffield}
  \institution{\textsuperscript{3}School of Computing \& Digital Technologies, Sheffield Hallam University}
  \institution{\textsuperscript{4}Centre for Robotics \& AI, University of Manchester}
  \country{United Kingdom}}

\renewcommand{\shortauthors}{Adikari et al.}

\begin{abstract}
Reinforcement Learning (RL) systems are typically trained using a single, well-specified scalar reward function. However, real-world decision-making tasks often involve multiple, competing objectives, such as performance versus efficiency, where ground-truth reward functions are difficult to specify or inaccessible. While Multi-Objective RL (MORL) addresses such trade-offs by modeling rewards as vectors, existing approaches typically assume access to a well-specified reward function for each objective, inheriting the same challenges faced by single-objective RL. Meanwhile, Preference-based RL (PbRL) has shown great potential in solving complex tasks without access to a pre-defined reward function through reward learning from feedback, yet has largely been studied in single-objective settings. In this work, we bridge this gap with LEMUR: Learning to Align with Multi-Objective Reinforcement Learning from Preference feedback, a novel framework where an agent interactively learns from the preferences of multiple teachers to learn optimal multi-objective policies. Our approach jointly learns policies and multiple objective-specific reward models from feedback, enabling agents to effectively balance competing objectives during learning. We evaluate LEMUR on a variety of benchmark multi-objective environments, and empirical results demonstrate its superior performance over existing baselines. Our method presents a promising direction for solving multi-objective decision-making tasks without pre-defined reward functions.
\end{abstract}


\newcommand{\BibTeX}{\rm B\kern-.05em{\sc i\kern-.025em b}\kern-.08em\TeX}

\begin{document}

\maketitle

\begingroup
\renewcommand{\thefootnote}{*}
\footnotetext{Corresponding author: manith.adikari@manchester.ac.uk}
\endgroup

\section{Introduction}

Reinforcement Learning (RL) has achieved remarkable success in training autonomous agents, from game-playing \cite{DBLP:journals/corr/MnihKSGAWR13} to robotics \cite{tansim}. By formalizing learning as the maximization of cumulative rewards through trial-and-error learning \cite{sutton_reinforcement_1998}, RL provides a natural framework for training autonomous `goal-seeking' agents \cite{McCarthy1997}. However, standard RL relies on two critical assumptions: that the goal can be represented by a single \textit{scalar reward}, and that this reward function is \textit{well-specified}. In practice, these assumptions rarely hold in complex, real-world domains.

Real-world tasks often involve \textit{multiple, competing} objectives \cite{dulac-arnold2019challenges}, such as balancing speed versus safety in autonomous driving \cite{wang2026learning}, or throughput versus energy efficiency in robotics \cite{huang2022constrained,kouritem2022multi}.

The field of Multi-Objective Reinforcement Learning (MORL) addresses this by modeling \textit{rewards as vectors} to find a set of Pareto-optimal policies \cite{roijers2013survey}: policies where no objective's expected returns can be increased without decreasing the returns of other objectives. However, existing MORL approaches typically assume that the ground-truth reward function for each objective is accessible and manually specified \cite{hayes_practical_2022}. Thus, they inherit the same challenges of reward specification from the single-objective RL domain, now across multiple objectives. Manually designing a reward function that can achieve an adequate balance between competing objectives is challenging and can lead to oversimplification, which results in suboptimal policies \cite{knox2012humans} and possible reward exploitation \cite{amodei2016concrete}. To avoid complicated reward engineering, Preference-based RL (PbRL) learns reward models directly from feedback \cite{christiano_deep_2017}, which leads to better alignment between the system's behavior and preferences. Despite progress in reward learning for single-objective RL, this problem remains largely unexplored in the MORL setting.  Work that combines preference feedback with multiple objectives either infers how a single user weights objectives whose rewards are known \cite{hwang_promptable_2023,hu2026humanintheloop}, learns a user-conditioned reward to personalize behavior to one user at a time \cite{poddar2024personalizing}, or targets Large Language Model (LLM) post-training \cite{bakker2022fine,rame2023rewarded,yang2024rewardsincontext}. These approaches do not jointly learn separate objective-specific reward models from multiple teachers online while recovering a set of policies that trade those objectives off. 

Reward specification is already a major bottleneck in scaling single-objective RL settings, and becomes even more critical in MORL. Two key challenges arise. First, the agent must optimize multiple objectives whose underlying reward functions are complex, implicit, or inaccessible. Second, even when feedback is available, collapsing multi-criteria reward signals into a single scalar obscures the very trade-off structure that MORL is designed to optimize \cite{vamplew2011empirical, roijers2013survey, pmlr-v235-sorensen24a}. Consider training a robotic system: \textit{non-expert} operators can reliably judge task-level success, such as whether objects were grasped or placed correctly, while expert operators are required to assess fine-grained criteria such as grasp stability, long-term wear, or safety-related objectives. Collapsing such \textit{heterogeneous} feedback into a single reward risks conflating \textit{distinct} objectives and diluting expert signals, motivating the need for \textit{separate}, objective-specific reward models learned from the appropriate source of feedback.

This creates a crucial gap: \textit{How can agents learn optimal trade-offs between conflicting objectives when the reward functions are unknown and must be inferred from feedback?}

To address this gap, we introduce \textbf{LEMUR}: \textbf{Le}arning to Align with \textbf{Mu}lti-Objective \textbf{R}einforcement Learning from Preference Feedback, a framework for learning to balance multiple objectives without pre-defined reward functions. Our approach jointly learns policies and multiple objective-specific reward models from preference feedback, enabling agents to effectively balance competing objectives during learning. By removing the assumption of known reward functions and explicitly modeling multiple objectives, our method tackles a key challenge towards scaling and aligning MORL in real-world domains.
To summarize, our main contributions are: 
\begin{itemize}
    \item We propose a novel framework, LEMUR, which learns \emph{multiple} objective-specific reward models from preference feedback and then optimizes policies against these reward models using multi-objective reinforcement learning. This enables agents to effectively solve multi-objective decision-making tasks without access to pre-defined reward functions, while naturally accommodating heterogeneous sources of feedback, where different teachers may hold expertise over different objectives. 
    \item Our extensive experiments demonstrate that LEMUR outperforms baselines across a range of benchmark multi-objective environments, namely high-dimensional continuous control tasks,  and further show LEMUR's robustness to label noise, constrained feedback budgets, and scaling to more objectives.
\end{itemize}

\begin{figure*}[t]
    \centering    \includegraphics[width=1.0\linewidth]{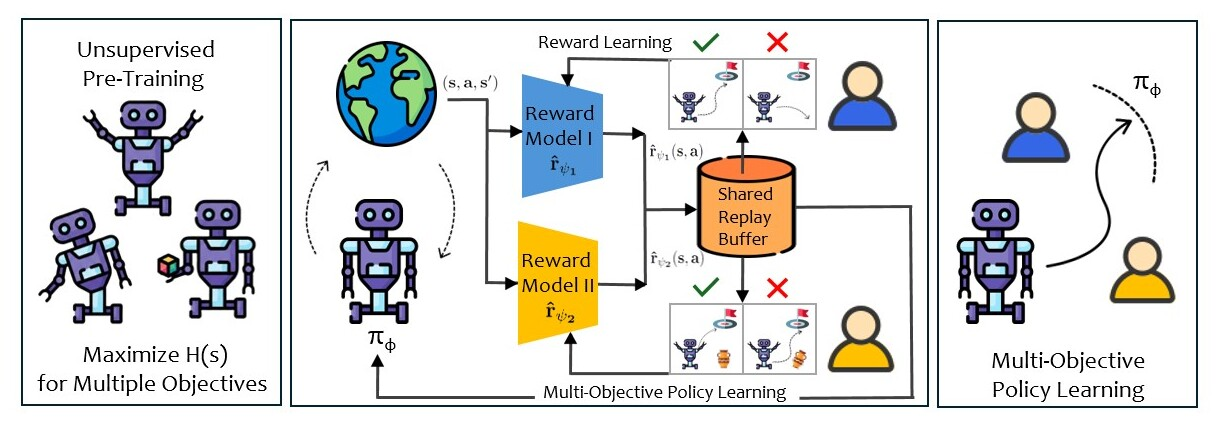}
    \caption{Illustration of our framework LEMUR: (1) Unsupervised Pre-training for the MORL agent to explore and collect diverse experiences via maximizing state entropy H(s). (2) Reward learning from Preference feedback, where each reward model is learned separately from the preferences queried from each teacher. The reward models are used to dynamically relabel the state-action pairs as a reward vector for each objective (i.e., each teacher's preferences).  (3) Multi-Objective RL agent denoted by $\pi_{\phi}$ uses each of the trained reward models to do multi-objective policy optimization to maximize the expected vector rewards.}
    \label{fig:LEMUR} 
\end{figure*}

\section{Preliminaries}
\label{prelims}
\textbf{Multi-Objective Reinforcement Learning.} We formulate the problem as a Multi-Objective Markov Decision Process (MOMDP)~\cite{white1982multi}, defined by the tuple $\langle \mathcal{S}, \mathcal{A}, \mathcal{T}, \gamma, \mathbf{r} \rangle$. Here, $\mathcal{S}$ and $\mathcal{A}$ denote state and action spaces respectively, $\mathcal{T}$ the transition dynamics, and $\gamma \in [0, 1)$ the discount factor. Unlike standard RL, the reward is a vector $\mathbf{r}(s, a) \in \mathbb{R}^d$ comprising $d$ distinct objectives. The agent's goal is to maximize the expected discounted vector returns $J(\pi) = \mathbb{E}_{\pi} [ \sum_{t=0}^{\infty} \gamma^t \mathbf{r}(\mathbf{s}_t, \mathbf{a}_t) ]$. A policy $\pi : \mathcal{S} \to \Delta(\mathcal{A})$ maps a state $s \in \mathcal{S}$ to a probability distribution over actions, where $\pi(a \mid s) = \mathbb{P}(A_t = a \mid S_t = s)$ gives the probability of choosing action $a \in \mathcal{A}$. Since no single policy typically maximizes all objectives in MORL, the agent learns a set of policies $\Pi$ representing optimal trade-offs ~\cite{hayes_practical_2022}.

\textbf{Soft Actor-Critic (SAC).} SAC~\cite{haarnoja2018soft} is an off-policy actor-critic algorithm grounded in the maximum entropy framework. It augments standard single-objective RL with an entropy term to encourage exploration. The agent aims to maximize $J(\pi) = \mathbb{E}_{\pi} [\sum_{t} \gamma^t (r_t + \alpha \mathcal{H}(\pi(\cdot|\mathbf{s}_t)))]$.

\textbf{Reward Learning from Preference Feedback.}
We follow the standard Preference-based RL (PbRL) framework, which uses `latent rewards' as a proxy for values~\cite{christiano_deep_2017}. A teacher (e.g., a human) provides preferences between the two agent behaviors, and the agent then uses this feedback to learn a reward function \cite{lee2021bpref}. Each teacher's reward function is learned independently using their respective preference feedback. Similar to prior work, we use preferences over \textit{pairs} of trajectory segments, $(\sigma^0, \sigma^1)$. The teacher provides a label $y$ to indicate their preference. To learn a reward function $\hat{r}$ parameterized by $\psi$, we employ the Bradley-Terry model~\cite{bradley1952rank}, modeling the preference probability as $P[\sigma^1 \succ \sigma^0; \psi] = \frac{\exp \sum \hat{r}(\sigma^1)}{\exp \sum \hat{r}(\sigma^1) + \exp \sum \hat{r}(\sigma^0)}$. Given a dataset $\mathcal{D}$, the reward function is trained by minimizing the cross-entropy loss:
\begin{equation}
\label{preflearningloss}
    \mathcal{L}^{CE}(\psi) = - \mathbb{E}_{\mathcal{D}} \Big[ (1-y) \log P[\sigma^0 \succ \sigma^1] + y \log P[\sigma^1 \succ \sigma^0] \Big].
\end{equation}
The learned reward function can then be used to update the policy with an RL algorithm to maximize expected returns.
\section{Problem Setup}
\label{sec:problem_setup}
In this section, we present our formulation of Multi-Objective Reinforcement Learning (MORL) \textit{without pre-defined} reward functions for \textit{multiple, conflicting} objectives. 

\paragraph{Latent Reward Vector.} In standard MORL, the environment exposes a known reward vector $\mathbf{r}(s, a) \in \mathbb{R}^{d}$ over $d$ components. In our work, the agent \textit{does not} have access to it. Instead, we assume $m$ distinct, conflicting teachers, where each teacher $j$ holds a preference profile $\boldsymbol{\omega}_j \in \Delta^{d}$. 
Formally, teacher $j$'s latent utility is the scalarization of the native reward under the teacher's profile,
\begin{equation}
\label{eq:teacherutility}
    u_j(s,a) = \boldsymbol{\omega}_j^\top \mathbf{r}(s,a),
\end{equation}
and its preference over a trajectory segment is governed by the corresponding segment utilities $U_j(\sigma) = \sum_{t} u_j(s_t, a_t)$. Neither $\mathbf{r}$ nor $u_j$ is observed by the agent: the teacher's feedback is the only evidence available, and $\hat{r}_{\psi_j}$ is an estimate of $u_j$ recovered from it.
Since these utilities are inaccessible, the agent must approximate them. We then define a parameterized reward vector $\mathbf{\hat{r}}_\psi(s, a) = [\hat{r}_{\psi_1}(s, a), \dots, \hat{r}_{\psi_m}(s, a)]^\top \in \mathbb{R}^{m}$, where each component is a reward model learned from each teacher $j$'s preference feedback \cite{christiano_deep_2017}. Note that $d$ is a property of the environment and $m$ is the number of teachers. For example, $d=3$ might represent the rewards for [forward velocity, jump height, energy cost], over which $m=2$ teachers hold different preferences; the agent therefore learns a two-dimensional reward vector over a three-dimensional native reward.

\paragraph{Multi-Objective RL Optimization.}
The agent's goal is to maximize the expected discounted vector returns. We adopt the utility-based approach \cite{roijers2013survey}, the most prevalent MORL formalism, defining a scalarization function $f_{\mathbf{w}}(\hat{\mathbf{r}}_\psi) = \mathbf{w}^\top \hat{\mathbf{r}}_\psi$, where $\mathbf{w} \in \Delta^{m}$ is a weight vector on the simplex (i.e.\ $\sum_i w_i = 1$) optimizing over the $m$ teachers' rewards. We therefore define an optimal MORL agent as the policies belonging to the Convex Coverage Set (CCS), the subset of the Pareto front, optimal for linearly scalarized preferences \cite{roijers2013survey}.

To summarize, the MORL agent learns these policies by maximizing the expected returns of the \textit{learned} latent reward vector linearly scalarized by the weight ${w}$, which determines the trade-offs between objectives: 
\begin{equation}
    \label{lemurobjective}
    J(\pi) = \mathbb{E}_{\pi} \left[ \sum_{t=0}^{\infty} \gamma^t \mathbf{{w}}^T\mathbf{\hat{r}_\psi}(s_t, a_t) \right].
\end{equation}

\section{LEMUR}
\textbf{LEMUR} (\textbf{Le}arning to Align with \textbf{Mu}lti-Objective \textbf{R}einforcement Learning from Preference Feedback), illustrated in Figure \ref{fig:LEMUR} proceeds in three stages: (1) unsupervised pre-training, where the agent explores via intrinsic rewards to collect diverse experiences (Section \ref{unsupervisedtraining}); (2) reward learning, where multiple teachers are queried for preference feedback to train objective-specific reward models (Section \ref{sec:LEMURrewardlearning}); and (3) multi-objective RL training against the learned reward models (Section \ref{sec:LEMURpolicytraining}). Stages 2 and 3 repeat, continually improving both the reward models and the multi-objective policies. 

\subsection{Unsupervised Pre-training} 
\label{unsupervisedtraining}
Standard PbRL suffers from uninformative queries caused by the limited coverage of random initialization. To generate informative queries, LEMUR employs an unsupervised pre-training phase driven by intrinsic motivation~\cite{pmlr-v139-lee21i}. We encourage exploration by maximizing state entropy, approximated via a particle-based $k$-nearest neighbors ($k$-NN) estimator \cite{liu2021behavior}. The intrinsic reward $r^{\text{int}}(s_t) = \log(\|s_t - s_t^k\|)$ is the distance to the $k$-th nearest neighbor of $s_t$ in $\mathcal{B}_0$, and the explorer policy $\pi_\phi$ maximizes $\mathcal{J}_{\text{int}}(\phi) = \mathbb{E}_{\pi_\phi} \bigl[ \sum_{t=0}^{T} \gamma^t r^{\text{int}}(s_t) \bigr]$. This populates the buffer $\mathcal{B}_0$ with diverse behaviors, accelerating the subsequent multi-objective reward learning.

\begin{algorithm}[htbp]
\caption{LEMUR}
\label{alg:LEMUR}
\begin{algorithmic}[1]
\Require Feedback frequency $K$, queries per session $M$, teacher profiles $\{\boldsymbol{\omega}_j\}_{j=1}^{m}$, concentration $\kappa$
\State Initialize policies $\{\pi_i\}_{i=1}^{N}$ with weights $\{\mathbf{w}_i\}_{i=1}^{N}$
\State Initialize reward models $\{\hat{\mathbf{R}}_{\psi_j}\}_{j=1}^{m}$, datasets $\mathcal{D}_j \leftarrow \emptyset$
\State $\mathcal{B}_0 \leftarrow \textsc{Explore}(r^{int})$; $\mathcal{B} \leftarrow \mathcal{B}_0$ \Comment{Pre-training}
\For{each iteration}
    \If{iteration $\bmod K = 0$} \Comment{Reward learning}
        \For{each teacher $j = 1,\dots,m$}
            \For{$n = 1,\dots,M$}
                \State Sample $(\sigma^0,\sigma^1) \sim \mathcal{B}_0$ and $\boldsymbol{\omega}^{(n)} \sim \mathrm{Dir}(\kappa\,\boldsymbol{\omega}_j)$
                \State Query teacher $j$ under $\boldsymbol{\omega}^{(n)}$ for label $y^{(n)}$
                \State $\mathcal{D}_j \leftarrow \mathcal{D}_j \cup \{(\sigma^0,\sigma^1,\boldsymbol{\omega}^{(n)},y^{(n)})\}$
            \EndFor
            \State Update $\hat{\mathbf{R}}_{\psi_j}$ on $\mathcal{D}_j$ via Eq.~\ref{preflearningloss} under Eq.~\ref{eq:wcbt}
        \EndFor
        \State Relabel $\mathcal{B}$ with $\hat{r}_{\psi_j} = \boldsymbol{\omega}_j^\top \hat{\mathbf{R}}_{\psi_j}$ for all $j$
    \EndIf
    \State Collect transitions with $\{\pi_i\}$ into $\mathcal{B}$ \Comment{Policy learning}
    \For{each policy $i$ and gradient step}
        \State Sample $(s,a,\hat{\mathbf{r}}_\psi,s') \sim \mathcal{B}$; update $\pi_i$ on $\mathbf{w}_i^\top \hat{\mathbf{r}}_\psi$
    \EndFor
    \State Adapt $\{\mathbf{w}_i\}$ with PSA
\EndFor
\end{algorithmic}
\end{algorithm}
\subsection{Reward learning of Multiple Objectives from Preferences} 
\label{sec:LEMURrewardlearning}
A core challenge in our setup is that the agent does not have access to the ground-truth rewards, and the $m$ conflicting objectives are characterized instead by the conflicting preferences of $m$ teachers. Following the PbRL formulation in Section \ref{prelims} \& \ref{sec:problem_setup}, LEMUR learns a separate reward model $\hat{\mathbf{R}}_{\psi_j}(s, a)$ per teacher, trained by minimizing the cross-entropy loss between the model's predictions and that teacher's labels (Equation \ref{preflearningloss}).

\textbf{Weight-Conditioned Reward Models.} Recall that teacher $j$'s profile $\boldsymbol{\omega}_j$ encodes how it trades off the $d$ native reward components, giving utility $u_j=\boldsymbol{\omega}_j^\top\mathbf{r}$ (Equation~\ref{eq:teacherutility}). Since $\mathbf{r}$ is hidden, we learn it: following \cite{mu2025preference}, teacher $j$'s reward model $\hat{\mathbf{R}}_{\psi_j}: \mathcal{S}\times\mathcal{A} \to \mathbb{R}^{d}$ is approximated by a lightweight MLP that predicts the full component vector. The Bradley-Terry model compares \emph{scalar} segment scores, so the vector output is scalarized by a weight $\boldsymbol{\omega}$ and the preference likelihood is conditioned on it:
\begin{equation}
\label{eq:wcbt}
P_{\psi_j}\!\left[\sigma^1 \succ \sigma^0 \mid \boldsymbol{\omega}\right] = \frac{\exp \sum_t \boldsymbol{\omega}^\top \hat{\mathbf{R}}_{\psi_j}(s^1_t, a^1_t)}{\sum_{i\in\{0,1\}} \exp \sum_t \boldsymbol{\omega}^\top \hat{\mathbf{R}}_{\psi_j}(s^i_t, a^i_t)}.
\end{equation}
Each label is thus a binary judgement of which segment is better \emph{under one specific trade-off} $\boldsymbol{\omega}$.

A label constrains only one linear combination $\boldsymbol{\omega}^\top\hat{\mathbf{R}}_{\psi_j}$ of the $d$ components; querying always at $\boldsymbol{\omega}_j$ would leave the other $d{-}1$ directions unconstrained by the labels. We therefore issue the $n$-th query to teacher $j$ under a \emph{query weight} $\boldsymbol{\omega}^{(n)} \sim \mathrm{Dir}(\kappa\,\boldsymbol{\omega}_j)$ (Dirichlet distribution) to prompt a slightly different trade-off over the same $d$ components, centred on $\boldsymbol{\omega}_j$ (the Dirichlet mean) with spread controlled by $\kappa$ (larger $\kappa$ stays closer to $\boldsymbol{\omega}_j$, and $\kappa\!\to\!\infty$ recovers a standard Bradley-Terry reward on $u_j$. We refer to prior work on this \cite{mu2025preference,zhu2023scaling,kim2025conflictaverse} and Appendix~\ref{app:kappa} for further reference. 

Each $\psi_j$ is trained by minimizing the cross-entropy loss (Equation~\ref{preflearningloss}), with the likelihood of Equation~\ref{eq:wcbt} evaluated at each label's own query weight, on teacher $j$'s data alone; the $m$ models share no parameters. Teacher $j$'s scalar reward is the projection onto its own profile, $\hat{r}_{\psi_j}(s,a) = \boldsymbol{\omega}_j^\top \hat{\mathbf{R}}_{\psi_j}(s,a)$, an estimate of its utility $u_j$. Stacking the $m$ scalars gives the learned reward vector $\hat{\mathbf{r}}_\psi \in \mathbb{R}^{m}$, which the MORL agent scalarizes under its own weight $\mathbf{w} \in \Delta^{m}$ (Equation~\ref{lemurobjective}). The two weights act at different levels: $\boldsymbol{\omega}_j$ combines the $d$ components into one teacher's utility, while $\mathbf{w}$ trades off the $m$ teachers against one another.

The agent never observes $\mathbf{r}$ or any $u_j$, and this follows prior preference-based MORL \cite{mu2025preference}. For fair comparison, we ensure all baselines employ the same method. Additionally, we show that LEMUR still learns effective compromise policies (Appendix~\ref{app:wcvsensemble}) with a standard reward model used in single-objective PbRL (an ensemble of unconditioned MLPs per teacher) \cite{lee2021bpref,pmlr-v139-lee21i}. 

\textbf{Query Selection.} Which segment pairs are shown is a separate choice from the query weight. As in standard PbRL, pairs are sampled uniformly from $\mathcal{B}_0$, giving broad coverage with no extra machinery; an entropy-based active sampler \cite{lee2021bpref} yields only a modest gain (Appendix~\ref{app:query_sampling}).

\subsection{Multi-Objective RL Training}
\label{sec:LEMURpolicytraining} 
Given the parameterized reward vector $\mathbf{\hat{r}}_\psi(s, a)$, LEMUR trains the MORL agent to maximize expected latent \textit{vector} rewards (Equation \ref{lemurobjective}). For policy optimization, we leverage MORL/D, a state-of-the-art Multi-Objective Soft Actor-Critic (MO-SAC) algorithm \cite{felten_multi-objective_2024} which learns a set of independent SAC policies and applies an evolutionary strategy for policy search. This off-policy choice is deliberate: reusing past experience from the replay buffer is essential for sample efficiency under a limited feedback budget.

\textbf{Policy Set and Weight Adaptation.} LEMUR does not infer a single compromise between teachers. Instead, MORL/D maintains a population of $N$ policies $\Pi = \{\pi_i\}_{i=1}^{N}$, each assigned its own scalarization weight $\mathbf{w}_i \in \Delta^{m}$ over the $m$ teachers and trained with SAC on $\mathbf{w}_i^\top \hat{\mathbf{r}}_\psi$ (Equation~\ref{lemurobjective}). The weights are initialized across the simplex and adapted during training by Pareto Simulated Annealing (PSA) \cite{felten_multi-objective_2024}, which moves each $\mathbf{w}_i$ in response to the current policies and their distance to non-dominated solutions, focusing training on feasible regions of the trade-off surface while maintaining diversity. These policy weights $\mathbf{w}_i$ are distinct from the teacher profiles $\boldsymbol{\omega}_j$: the former trade off teachers against one another and are never learned from feedback, while the latter define what each teacher prefers. Following the utility-based view of MORL \cite{roijers2013survey,hayes_practical_2022}, the output of LEMUR is the set $\Pi$, from which each stakeholder selects the policy that best serves their own preferences (Appendix~\ref{app:policyset}).
 
\textbf{Cooperation via Shared Buffer.} 
To facilitate information exchange across policies learning different trade-offs, all policies store and sample from a common replay buffer $\mathcal{B}$. This enables policies to learn from diverse experiences collected under different preference weightings, improving sample efficiency, a critical consideration given the limited feedback budget.

\textbf{Relabeling of Vector Rewards}. Combining off-policy RL with a reward function learned from preferences introduces non-stationarity: as the reward model is updated with new feedback, rewards associated with past transitions in the buffer become stale, destabilizing learning. To address this, LEMUR employs a vector reward relabeling strategy inspired by prior work \cite{pmlr-v139-lee21i}. After each reward-model update, all transitions stored in the buffer are relabeled with the current reward models. This ensures the agent always trains on updated rewards, synchronizing policy and reward learning while preserving the sample efficiency of our off-policy approach.

\section{Experiments}
\label{sec:experiments}
Our experiments address three questions: (1) Can LEMUR learn multi-objective policies that balance multiple \textit{learned} reward models from \textit{multiple} teachers? (2) How does LEMUR compare to existing baselines on multi-objective benchmarks? (3) Does explicitly learning multiple reward models for conflicting feedback outperform aggregating feedback into a single reward model? For all experiments, we report the mean across five random seeds with standard error. Additional implementation details are reported in Appendices \ref{app:lemur_implementation} \& \ref{app:implementationdetails}.
\begin{figure}[htbp]
  \centering
    \includegraphics[width=0.22\linewidth, height=0.22\linewidth]{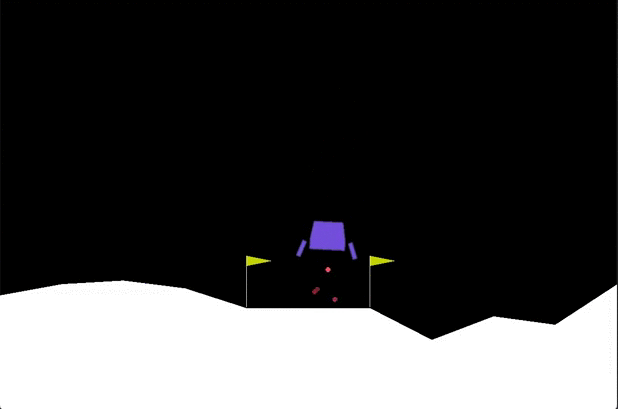}\hspace{0.08cm}%
  \includegraphics[width=0.22\linewidth, height=0.22\linewidth]{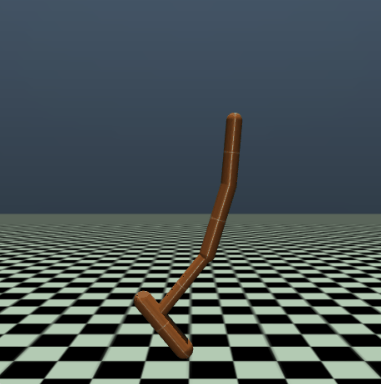}\hspace{0.08cm}%
  \includegraphics[width=0.22\linewidth, height=0.22\linewidth]{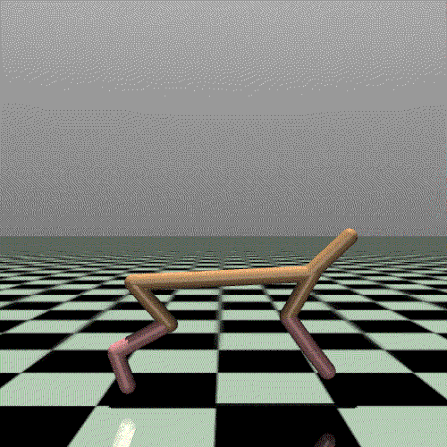}\hspace{0.08cm}%
  \includegraphics[width=0.22\linewidth, height=0.22\linewidth]{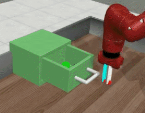}
  \caption{Benchmark environments evaluated in our experiments: \texttt{MO-LunarLander}, \texttt{MO-Hopper}, \texttt{MO-Cheetah}, and \texttt{MO-MetaWorld}.}
  \label{fig:environments}
\end{figure}
\begin{figure*}[t]
    \centering
    \includegraphics[width=1\linewidth]{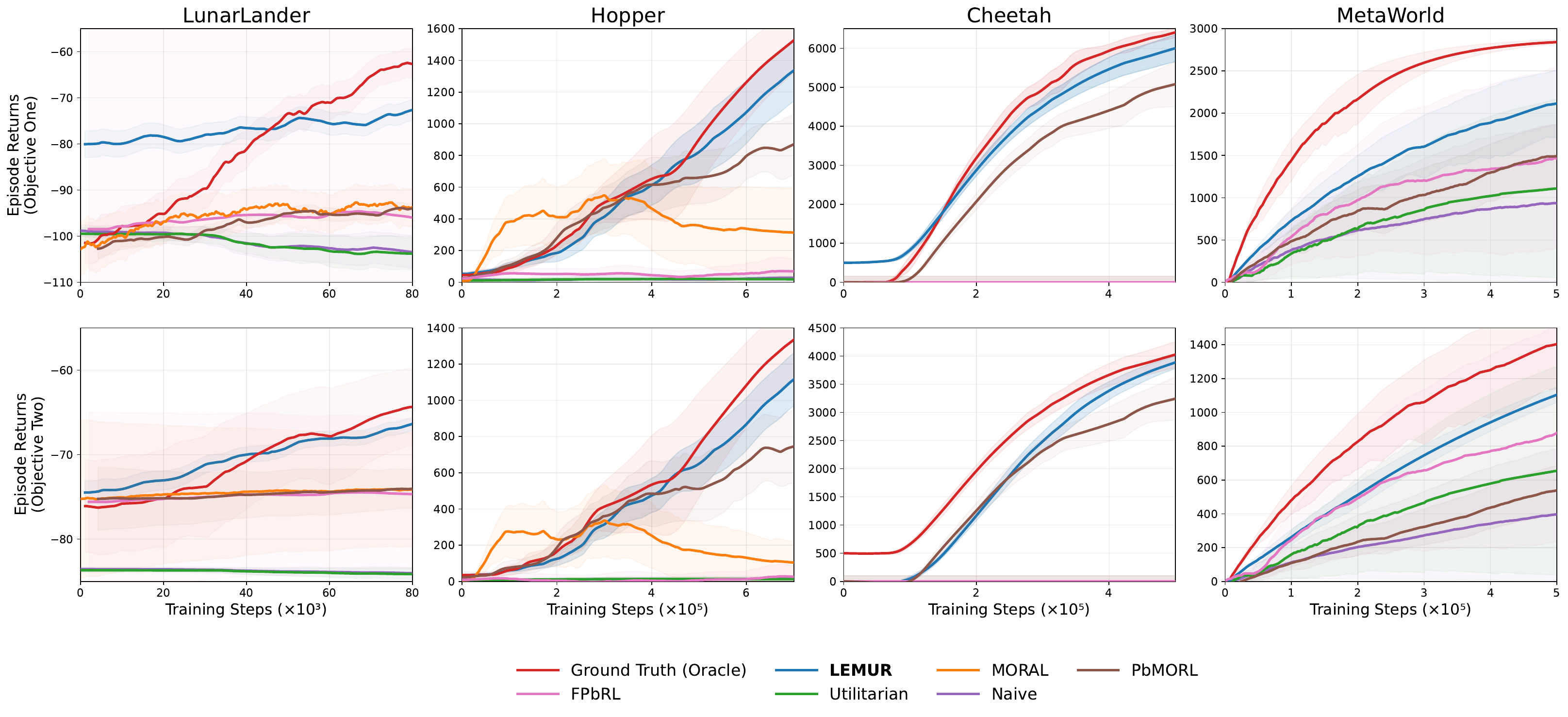}
    \caption{Learning curves on all benchmark environments: \texttt{MO-LunarLander}, \texttt{MO-Hopper}, \texttt{MO-Cheetah}, and \texttt{MO-MetaWorld}. Curves depict the true objective returns (inaccessible to the agent), averaged across five seeds, with shaded regions representing standard error. \textbf{LEMUR} (blue) most closely tracks the \textbf{Oracle} (red) on both objectives simultaneously. Baselines that aggregate conflicting feedback (\textbf{Naive}, \textbf{Utilitarian}) fail to make progress, while the external baselines (\textbf{MORAL}, \textbf{PbMORL}, \textbf{FPbRL}) learn but consistently trail LEMUR.}
    \label{fig:mainplots}
\end{figure*}

\textbf{Benchmark Environments \& Setup.}
We evaluate LEMUR on high-dimensional environments from the \textit{MORL-Generalization} benchmark \cite{teoh2025on}: \texttt{MO-LunarLander}, \texttt{MO-Hopper}, and \texttt{MO-Cheetah}. To test the framework beyond locomotion, we additionally construct \texttt{MO-MetaWorld}, a multi-objective adaptation of the `drawer-close' task from the Meta-World robot-manipulation suite \cite{yu2020meta}. Manipulation is a natural setting for multi-objective reward learning \cite{zhang2024graspxl,torne2026freeform,hwang_promptable_2023}, since a single scalar reward conflates competing criteria such as task success, motion smoothness, and energy cost; as Meta-World tasks are natively single-objective, we pair the dense task-progress reward with a control-effort penalty. Full details of environments are given in Appendix~\ref{app:envdetail}.

\textbf{Teachers \& Feedback Protocol}. Following standard benchmarking in PbRL \cite{lee2021bpref,christiano_deep_2017}, we use scripted teachers that generate feedback according to the components of the ground-truth vector reward, enabling quantitative evaluation. The ground-truth rewards remain inaccessible to the agent, which must jointly learn the conflicting preferences and optimize to find a balance. Our main experiments use two conflicting teachers, the fundamental version of the problem, and we scale to more teachers in Figure~\ref{fig:ablation_teachers_scaling}. In \texttt{MO-LunarLander}, Teacher A rewards precise, stable landings while Teacher B prioritizes fuel conservation; in \texttt{MO-Hopper} and \texttt{MO-Cheetah}, Teacher A prefers fast locomotion while Teacher B prefers slow, energy-efficient gaits; and in \texttt{MO-MetaWorld}, Teacher A rewards task progress while Teacher B rewards smooth, low-effort actuation. Following PbRL benchmarking protocol \cite{lee2021bpref}, we relax the assumption of noiseless teachers by flipping a fraction of labels to emulate teacher error (Section~\ref{sec:ablations}), and we further validate LEMUR with real human feedback in Section~\ref{app:human}. Full details of teacher profiles are given in Appendix~\ref{app:envdetail}, and we show that LEMUR remains close to the Oracle as teacher conflict increases (Appendix~\ref{app:conflict_levels}).

\textbf{Baselines.}
We compare against five baselines spanning distinct strategies for preference aggregation and learning from multiple objectives: (1) a \textbf{Utilitarian} agent, a single SAC agent optimizing the arithmetic mean of the independently learned rewards; (2) \textbf{Naive} data pooling, which trains one monolithic reward model on all conflicting feedback, akin to standard Reinforcement Learning from Human Feedback (RLHF); (3) \textbf{MORAL}~\cite{peschl_moral_2021}, which recovers per-teacher rewards via Adversarial Inverse Reinforcement Learning (AIRL) and learns a scalarization over them; (4) \textbf{PbMORL}~\cite{mu2025preference}, a recent preference-based multi-objective method learning a weight-conditioned vector reward from pairwise feedback; and (5) \textbf{FPbRL}~\cite{siddique2023fairness}, which aggregates learned per-teacher rewards through a Generalized Gini Welfare scalarization to optimize for fairness. We additionally report (6) an \textbf{Oracle}, the same agent trained on the teachers' ground-truth utilities under the same budget, as a reference for achievable performance.

Unless otherwise noted, all baselines share LEMUR's interactive loop, pre-training buffer, reward-model architecture, teacher profiles, query and environment-step budgets; Utilitarian, Naive, and PbMORL also use its weight-conditioned reward model \cite{mu2025preference}, while MORAL and FPbRL keep their own reward-learning mechanisms, their defining contributions. The primary distinction between methods is therefore \emph{how conflicting reward signals are aggregated and optimized} (Appendix~\ref{app:implementationdetails}).

\textbf{Evaluation Protocol.} Following standard PbRL practice \cite{christiano_deep_2017,pmlr-v139-lee21i,lee2021bpref}, we evaluate on ground-truth rewards, which are never available to the agent. For each teacher $j$, we record the episodic return of its true utility, $G_j(\tau) = \sum_{t} \boldsymbol{\omega}_j^\top \mathbf{r}(s_t, a_t)$. Following standard MORL benchmarking practice \cite{hayes_practical_2022,felten_multi-objective_2024,teoh2025on}, a multi-policy method is evaluated by the utility each user obtains from its best-serving policy in the learned set; we therefore report, for each teacher, the best return across the policy set over recent episodes. PbMORL uses the same six-policy population as LEMUR, so this comparison is matched, while single-policy baselines are evaluated identically over their one policy. Full details are given in Appendix~\ref{app:evalprotocol}.

\subsection{Results \& Analysis}
\label{sec:results}

Figure \ref{fig:mainplots} presents the learning curves for all methods across the benchmark environments. Across every environment, LEMUR is the method that most closely tracks the Oracle on both objectives simultaneously, effectively recovering policies that balance multiple, conflicting objectives. The Utilitarian and Naive agents remain flat with suboptimal returns throughout, supporting our hypothesis that aggregating conflicting reward signals into a single scalar degrades performance in such multi-objective settings.
\begin{figure}[t]
    \centering
    \includegraphics[width=0.75\columnwidth]{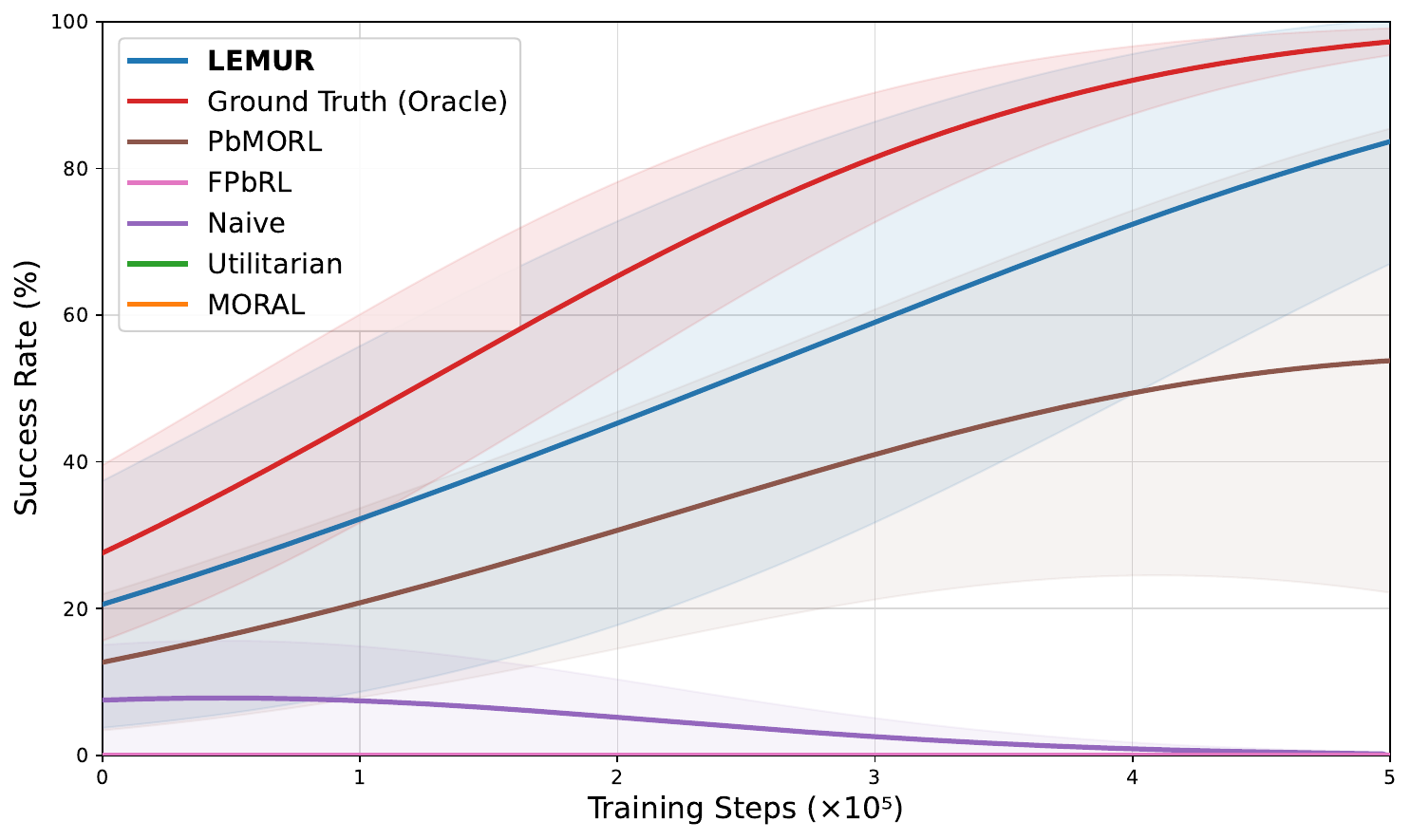}
    \caption{Task Success Rate (\%) learning curves on MetaWorld (Drawer Close Task).}
    \label{fig:lemur_metaworldsuccess}
\end{figure}

While MORAL improves early in MO-LunarLander and MO-Hopper, it fails to sustain this progress: in MO-Hopper its returns peak mid-training and then steadily decline. This is consistent with the known failure of out-of-distribution optimization due to static reward models; MORAL infers its per-teacher rewards offline from expert demonstrations via AIRL and holds them fixed. LEMUR instead does online policy optimization and addresses non-stationarity through \textit{vector reward relabeling}.

PbMORL and FPbRL, which both learn vector rewards from preference feedback, perform better than the aggregation baselines, yet still fall short of LEMUR. We attribute this to the assumptions both methods inherit. PbMORL trains a single weight-conditioned reward model over pooled feedback, implicitly assuming all preferences originate from one single teacher; under conflicting teachers the pooled model must average over conflicting labels, degrading the reward signal, most visibly in MO-Hopper and MO-Cheetah where it consistently trails LEMUR on both objectives. FPbRL preserves the vector structure but employs a fixed Generalized Gini welfare scalarization a priori, converging to a single welfare-optimal policy rather than a set of trade-offs: it achieves reasonable returns on MetaWorld, but neither achieves task success (Figure~\ref{fig:lemur_metaworldsuccess}) nor makes progress on the other environments. LEMUR reaches closest to the Oracle in task success rate.

\textbf{Multi-Objective Metrics}. We evaluate LEMUR using standard multi-objective metrics~\cite{hayes_practical_2022}. \textit{Hypervolume} (HV) measures the volume of objective space, rewarding policies that are both high-performing and broadly spread \cite{teoh2025on}; \textit{Sparsity} (SPS) measures the average distance between policies along the front, with lower values indicating more uniform coverage \cite{teoh2025on}. 

\begin{table}[htbp]
  \centering
  \small
  \setlength{\tabcolsep}{4pt}
  \resizebox{\columnwidth}{!}{%
    \begin{tabular}{lcccc}
      \toprule
      \multirow{2}{*}{\textbf{Environment}} & \multicolumn{2}{c}{\textbf{Hypervolume ($\uparrow$)}} & \multicolumn{2}{c}{\textbf{Sparsity ($\downarrow$)}} \\
      \cmidrule(lr){2-3} \cmidrule(lr){4-5}
      & \textbf{LEMUR} & \textbf{PbMORL} & \textbf{LEMUR} & \textbf{PbMORL} \\
      \midrule
      MO-LunarLander & $\mathbf{1.10\times10^4 \pm 3.48\times10^2}$ & $1.09\times10^4 \pm 5.74\times10^2$          & $\mathbf{3.1 \pm 2.7}$       & $4.1 \pm 1.4$ \\
      MO-Hopper      & $\mathbf{3.67\times10^6 \pm 2.84\times10^5}$ & $2.27\times10^6 \pm 3.11\times10^4$          & $\mathbf{250.1 \pm 5.8}$     & $1067.6 \pm 355.3$ \\
      MO-Cheetah     & $4.86\times10^7 \pm 1.34\times10^6$          & $\mathbf{4.90\times10^7 \pm 1.18\times10^6}$ & $\mathbf{3001.8 \pm 1580.6}$ & $4128.7 \pm 2825.5$ \\
      MO-MetaWorld   & $2.40\times10^6 \pm 9.71\times10^5$          & $\mathbf{3.03\times10^6 \pm 2.25\times10^6}$ & $\mathbf{428.6 \pm 422.9}$   & $3814.4 \pm 3619.3$ \\
      \bottomrule
    \end{tabular}%
  }
  \caption{Hypervolume and Sparsity (mean $\pm$ standard deviation over seeds) for LEMUR and PbMORL. Full results are in Appendix~\ref{app:additionalresults}.}
  \label{tab:morl_metrics_lemur_pbmorl}
\end{table}

As summarized in Table~\ref{tab:morl_metrics_lemur_pbmorl}, LEMUR attains the highest or comparable Hypervolume across all four environments: it clearly outperforms PbMORL on MO-Hopper and remains within one standard deviation of PbMORL on the other three. LEMUR also attains lower mean sparsity than PbMORL on all four environments, most clearly on MO-Hopper, where the variance across seeds is small, indicating that it learns policies that are competitive in performance while being more uniformly distributed over the trade-off space.  Full results are reported in Appendix~\ref{app:additionalresults}.


\begin{figure*}[t]
    \centering
    \begin{minipage}{0.32\textwidth}
        \centering
        \includegraphics[width=\linewidth]{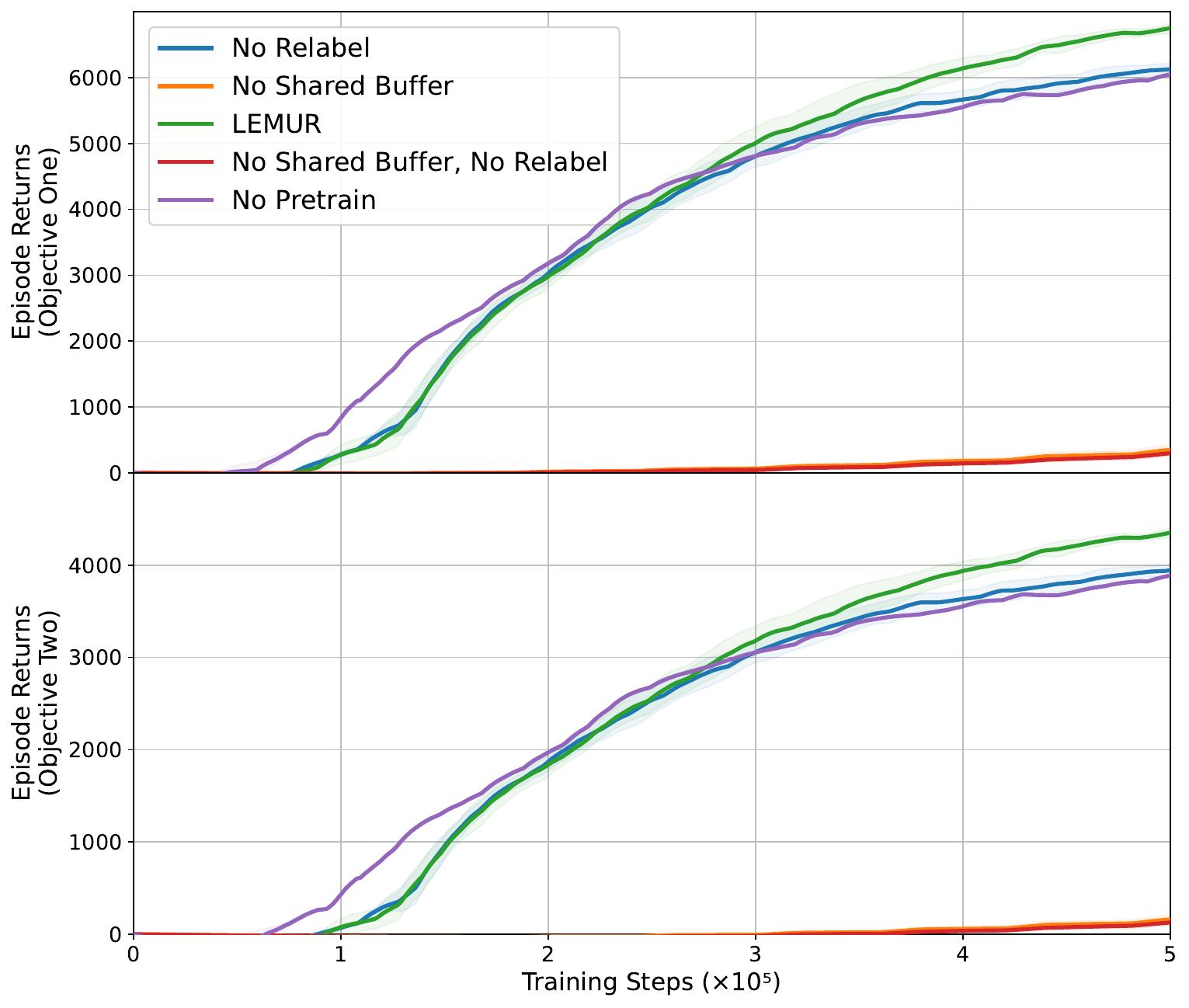}
        \vspace{0.1cm}
        \centerline{(a) Buffer Relabeling}
    \end{minipage}\hfill
    \begin{minipage}{0.32\textwidth}
        \centering
        \includegraphics[width=\linewidth]{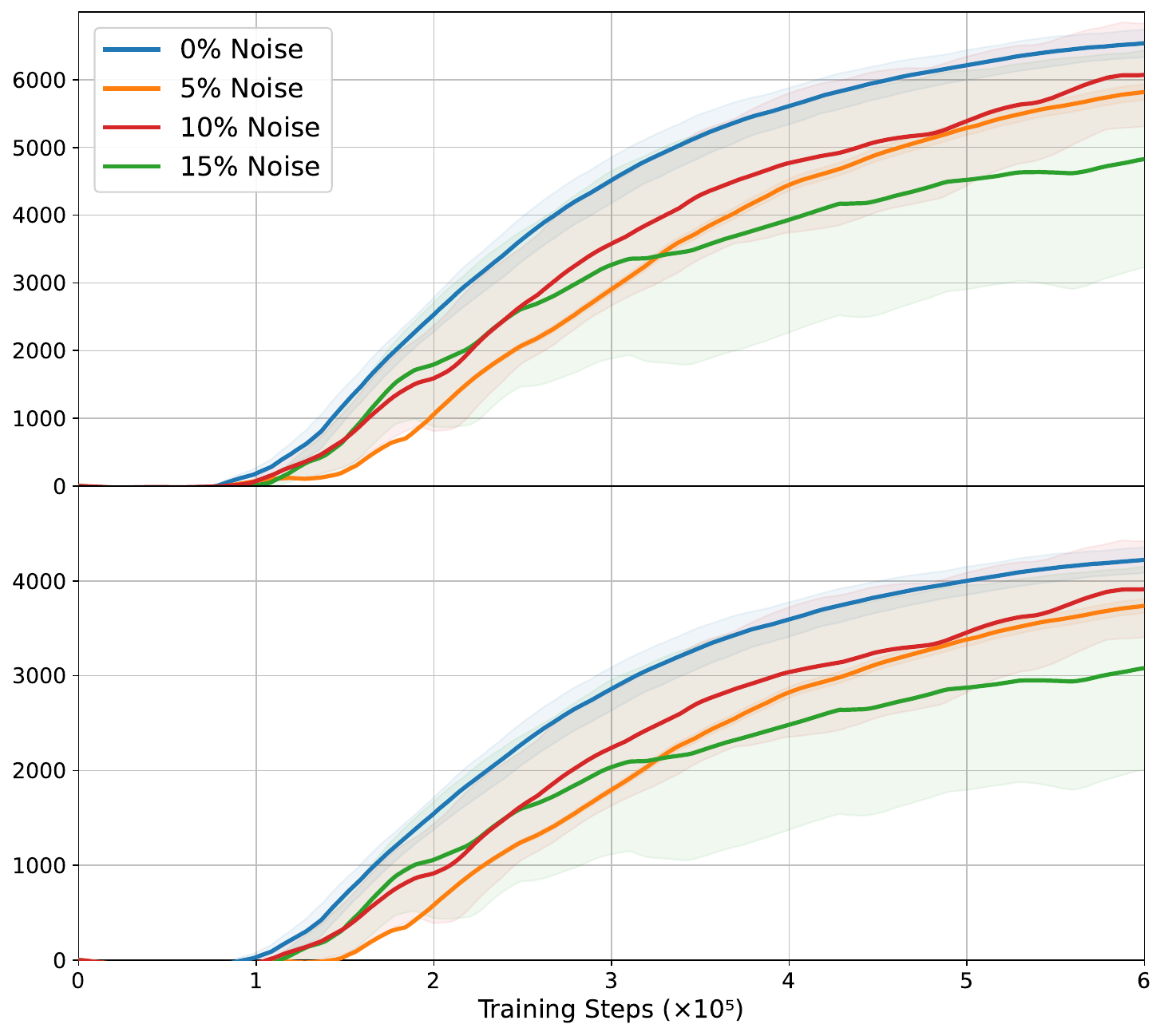}
        \vspace{0.1cm}
        \centerline{(b) Noisy Labels}
    \end{minipage}\hfill
    \begin{minipage}{0.32\textwidth}
        \centering
        \includegraphics[width=\linewidth]{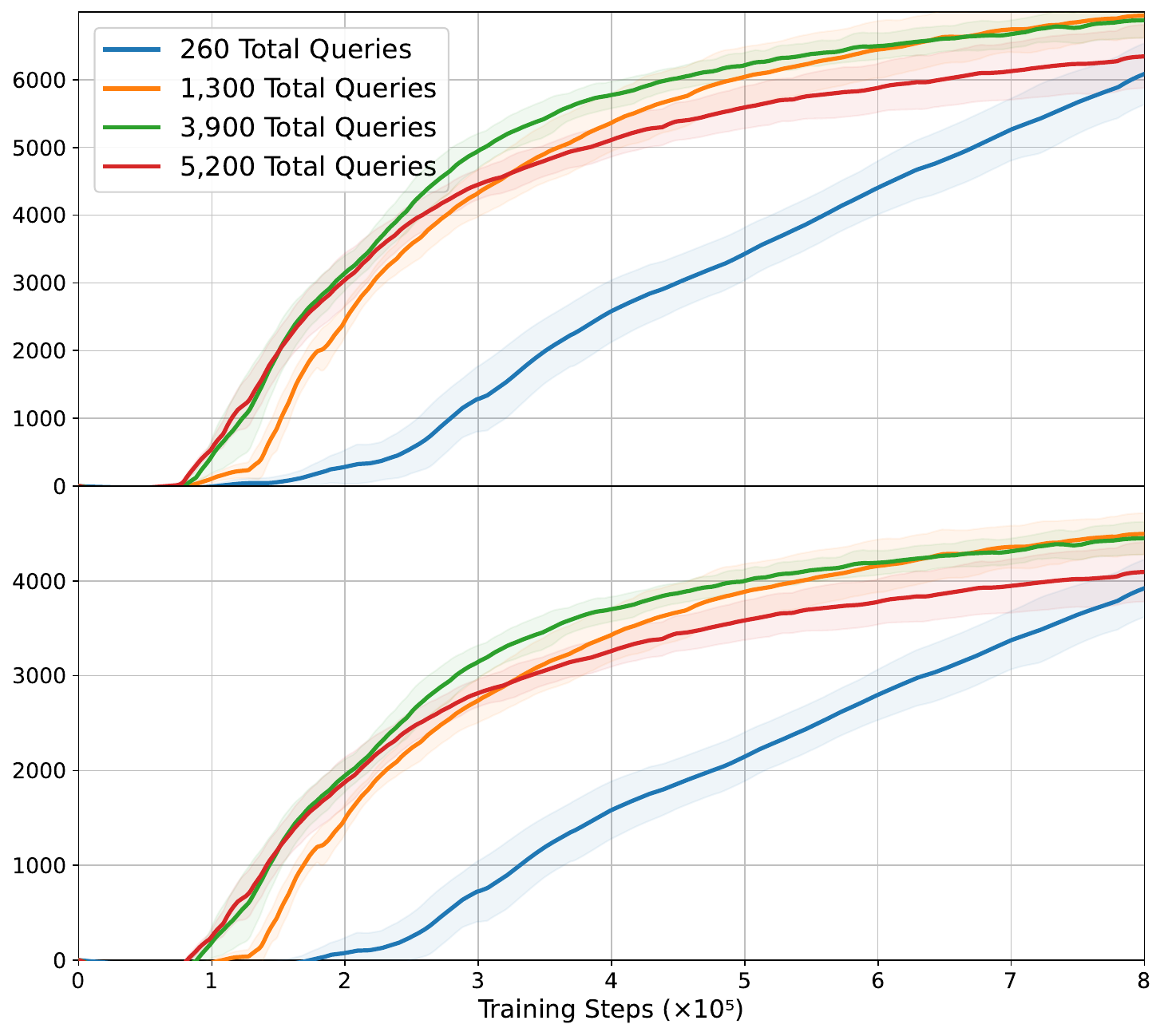}
        \vspace{0.1cm}
        \centerline{(c) Query Budget}
    \end{minipage}
    
    \caption{Ablation studies on \texttt{MO-Cheetah} evaluating the impact of (a) the shared buffer, vector reward relabeling, and unsupervised pre-training, (b) noisy teacher labels, and (c) varying the total query budget per teacher on agent returns for both objectives; by default LEMUR  uses 3900 queries (green). The results are averaged over multiple runs across five seeds.}
    \label{fig:three_ablations}
\end{figure*}

\textbf{Reward Model Alignment}. We also evaluate the learned reward models against the ground-truth teacher rewards \cite{lee2021bpref}, reporting \textit{Spearman} rank correlation over individual states and the \textit{Trajectory Alignment Coefficient} (TAC) \cite{muslimani2025towards} over whole trajectories. LEMUR's reward models align strongly on MO-Hopper ($\rho{=}0.945$, TAC $0.856$) and MO-Cheetah ($\rho{=}0.710$, TAC $0.898$), and moderately on MO-MetaWorld ($\rho{=}0.520$, TAC $0.347$), where they still exceed both baselines on these two metrics; full results are in Appendix~\ref{app:reward_metrics}.

\subsection{Ablation Studies}
\label{sec:ablations}
To validate the components of LEMUR, we conduct ablation studies on the high-dimensional \texttt{MO-Cheetah} domain; Figure \ref{fig:three_ablations} visualizes the learning curves.

\textbf{Impact of Shared Buffer, Relabeling, and Pre-training.} Figure \ref{fig:three_ablations}(a) isolates the contributions of the shared replay buffer, vector reward relabeling, and unsupervised pre-training. Disabling the shared buffer causes the most severe degradation. With the shared buffer intact, removing relabeling alone produces a modest drop relative to full LEMUR, confirming that recomputing rewards under the current models stabilizes learning against reward non-stationarity. Removing pre-training accelerates the earliest phase of training, but converges to lower final returns, indicating that the diverse initial buffer ultimately yields better reward models and policies.

\textbf{Robustness to Label Noise.} Figure \ref{fig:three_ablations}(b) corrupts a fraction of teacher labels (flipping preferences with probability up to 15\%), following PbRL benchmarking protocol \cite{lee2021bpref}. LEMUR is relatively robust to noise, with reasonable performance up to 15\%.

\textbf{Impact of Feedback Budget.} Figure \ref{fig:three_ablations}(c) varies the total query budget from 260 to 5,200 per teacher. Performance improves with budget, and larger budgets learn faster; notably, even 260 queries per teacher suffice to learn adequate policies on both objectives. This feedback efficiency is particularly important in preference-based RL, where queries are limited..

\begin{figure}[htbp]
    \centering
    \includegraphics[width=1\linewidth]{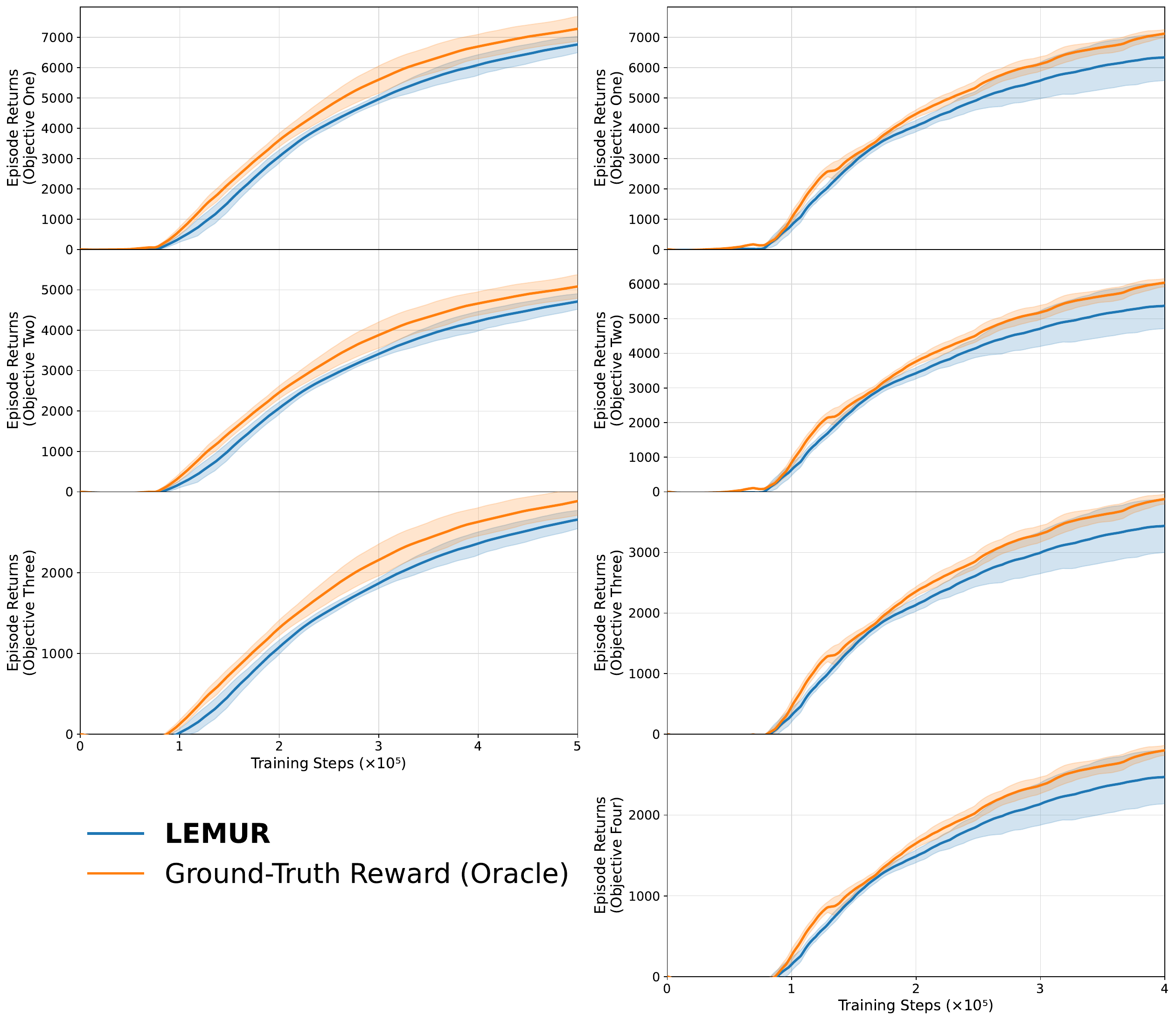}
    \caption{LEMUR scalability to higher-dimensional objective spaces, i.e. with more teachers. Episode returns for \textbf{(a)} 3-teachers and \textbf{(b)} 4-teachers, comparing policies trained with LEMUR (blue) versus ground-truth oracle rewards (orange).}
    \label{fig:ablation_teachers_scaling}
\end{figure}

\textbf{Scaling to More Objectives}. We next vary the teacher configuration on \texttt{MO-Cheetah}. Figure \ref{fig:ablation_teachers_scaling} extends LEMUR to three and four conflicting teachers: the learned-reward policies closely track the ground-truth oracle across all objectives, demonstrating that the per-teacher decomposition scales without modification, with each additional objective adding one reward model. 

\textbf{Additional Experiments.} Appendix~\ref{app:additionalexperiments} reports further experiments on \texttt{MO-Cheetah} and \texttt{MO-Hopper}. LEMUR remains close to the Oracle as teacher conflict increases and when teachers largely agree. Replacing the weight-conditioned reward model with an ensemble of unconditioned MLPs, the standard in single-objective PbRL \cite{lee2021bpref,pmlr-v139-lee21i}, removes the teacher profiles from the learner entirely, yet LEMUR still learns effective compromise policies (Appendix~\ref{app:wcvsensemble}). LEMUR is also compatible with non-linear scalarization, more robust to label noise than PbMORL, and benefits from off-policy optimization; query ablations show that segment length matters more than query selection.
\begin{figure*}[t]
    \centering
    \includegraphics[width=\textwidth]{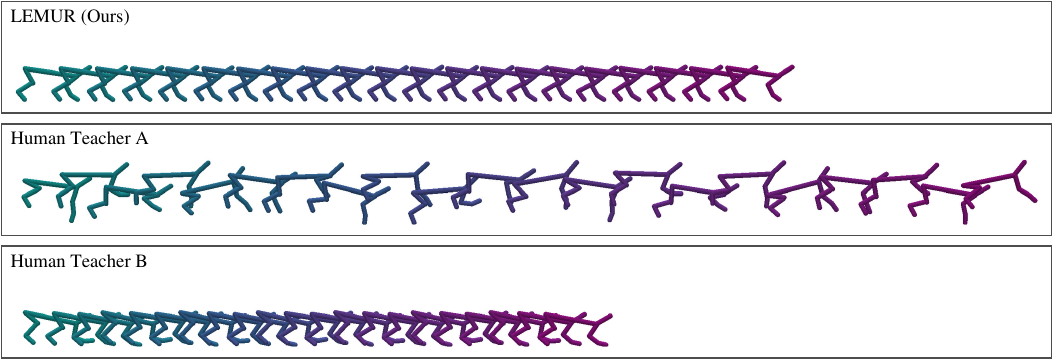}
   \caption{Trajectories on \texttt{MO-Cheetah} of policies trained from real human feedback: a policy from LEMUR (top), and single-objective SAC trained on each teacher's preferences alone. The agent's trajectory is shown over time on a shared spatial scale, so horizontal extent reflects distance traveled.}
    \label{fig:hitl_cheetah_trajectories}
\end{figure*}


\subsection{Human Experiment}
\label{app:human}
To validate LEMUR beyond scripted teachers, we train it on \texttt{MO-Cheetah} from the preferences of real human teachers. As in prior PbRL work with real human feedback, feedback was provided by the authors as a proof of concept \cite{christiano_deep_2017,pmlr-v139-lee21i}. Each teacher was given a conflicting objective through a short written instruction: Teacher A prefers fast forward locomotion, whereas Teacher B prefers calm, low-effort motion. Teachers labeled pairs of 50-step segments shown as videos, without access to the ground-truth rewards or the learned reward models, providing 200 queries each over 8 sessions in under 30 minutes per teacher (Appendix~\ref{app:humanprotocol}). As reference points, we train a single-objective SAC agent on each teacher's preferences alone.

\textbf{Qualitative Analysis.} Figure~\ref{fig:hitl_cheetah_trajectories} shows that real human feedback induces qualitatively distinct behaviors: the policy trained on Teacher A's preferences adopts long, bounding strides, whereas the policy trained on Teacher B's preferences adopts a low, crouched gait that covers less ground. LEMUR, trained jointly on both teachers' feedback, learns a set of policies from a single interactive run; the policy achieves balanced alignment with both teachers' behavioral preferences.

As in prior work \cite{christiano_deep_2017,pmlr-v139-lee21i,macglashan2017interactive}, our teachers are familiar with the task and method, so this experiment demonstrates feasibility with real human feedback rather than measuring the behavior of non-expert teachers. A larger-scale multi-teacher user study with independent participants is an important direction for future work.

\section{Related Work}
\label{sec:related_work}

\paragraph{\textbf{Reward Learning from Preference Feedback}.}
Designing reward functions is a primary bottleneck in scaling RL, as manual crafting is impractical and can induce unsafe behavior \cite{amodei2016concrete}; prior work instead learns rewards from demonstrations \cite{ng2000algorithms,abbeel2004apprenticeship}, language \cite{lin2022inferring}, scalar shaping feedback \cite{knox2009interactively,macglashan2017interactive}, or pairwise comparisons \cite{christiano_deep_2017}. Preference-based RL (PbRL) learns rewards from pairwise comparisons \cite{christiano_deep_2017,pmlr-v139-lee21i}; popularized as RLHF \cite{stiennon2020learning,ouyang2022training}, it typically trains a single reward model, aggregating diverse feedback into one scalar \cite{ouyang2022training} and failing to capture the multi-objective nature of human values \cite{pmlr-v235-sorensen24a}. Offline variants train rewards on fixed datasets before policy optimization \cite{shin_benchmarks_2023}, but static models suffer distribution shift as the policy diverges from the offline coverage, causing reward exploitation \cite{gao2023scaling,ye2024online}. Online, iterative RLHF mitigates this by collecting feedback alongside policy optimization \cite{dong2024rlhf,gao2023scaling,ye2024online}, which is more critical still in MORL, where the agent must span a space of diverse policies \cite{hayes_practical_2022}; offline MORL instead presupposes specified rewards or adequate dataset coverage \cite{yuan_moduli_2024,zhu2023scaling}. LEMUR circumvents this by jointly and interactively optimizing both the reward models and the multi-objective policy online.

\paragraph{\textbf{Multi-Objective Reinforcement Learning (MORL)}.}
MORL learns a set of policies approximating the Pareto frontier \cite{roijers2013survey,hayes_practical_2022}, via single-policy scalarization, weight-conditioned, or multi-policy methods; multi-task and Meta-RL are closely related \cite{chen2019meta,abdolmaleki2020distributional,sener2018multi}. Beyond linear scalarization, other work optimizes non-linear utilities of the return, including the expected scalarized return \cite{agarwal_multi-objective_2022,3709347.3743798}. Most of this literature assumes a vector of ground-truth reward functions \cite{hayes_practical_2022}, which is impractical in complex, real-world tasks. Interactive approaches elicit a user's preferences to select a trade-off, inferring scalarization weights from trajectory comparisons, demonstrations, or language \cite{hwang_promptable_2023,hu2026humanintheloop}, but still train on known objective rewards. LEMUR extends the MORL paradigm to the setting where the objectives must be inferred from preferences.

\paragraph{\textbf{Learning \& Alignment with Diverse Objectives}.}
Standard RLHF fails to capture the \textit{pluralistic} nature of human values \cite{pmlr-v235-sorensen24a}, and existing remedies rely on manual aggregation functions \cite{rodriguez2023multi}, expensive consensus datasets \cite{tessler_ai_2024}, static offline learning \cite{bakker2022fine}, or model heterogeneous feedback as hidden context without optimizing the trade-off between preferences \cite{siththaranjan2024distributional}. Outside of the language domain, variational preference learning \cite{poddar2024personalizing} learns a latent-conditioned reward from conflicting preferences, inferring each user's latent from a few labels to personalize behavior to that user; LEMUR instead keeps each teacher's objective explicit and recovers a set of policies trading teachers off against one another. In multi-objective multi-agent decision making, agents hold individual utilities over a shared vector reward \cite{ruadulescu2020multi}; LEMUR considers the converse, a single agent serving several stakeholders whose utilities must be learned. Our closest baselines learn rewards for multiple objectives but require expert demonstrations \cite{peschl_moral_2021}, assume a single teacher \cite{mu2025preference}, or require a priori welfare scalarization \cite{siddique_fairness_2023}.  LEMUR instead learns objective-specific reward models from separate feedback streams and jointly optimizes a multi-objective policy online via \textit{vector reward relabeling}, without offline reward-model training or expert demonstrations.

\section{Conclusion \& Limitations}
We propose LEMUR, a framework for Multi-Objective RL in domains where reward functions are unknown and must be inferred from the conflicting preferences of multiple teachers. LEMUR jointly learns the objectives and the policies that balance them, without expert demonstrations, pre-defined rewards, or a priori aggregation rules that existing methods require. Across multi-objective RL control and robotic manipulation benchmark environments, LEMUR outperforms aggregation baselines and recent preference-based multi-objective methods, and remains robust to label noise, reduced feedback budgets, and scaling to additional objectives. Several directions for future work are as follows. Our main evaluation uses scripted teachers, standard practice in PbRL for controlled and reproducible comparison \cite{lee2021bpref}. Our ablations highlight LEMUR's robustness, and our human experiment demonstrates feasibility with real feedback; a larger-scale study with independent participants is the natural next step. Finally,  active querying \cite{akrour2012april} offers a route to further reducing the number of queries, which is a promising path for deploying preference-based MORL in real-time.

\bibliographystyle{ACM-Reference-Format} 
\bibliography{sample}



\clearpage
\appendix 
\section*{Appendix}

\section{Extended Related Works}
\paragraph{Vector Rewards and the Limits of Scalar Reward.}
A key premise in modern RL is the \textit{reward hypothesis}: that goals can be adequately captured by maximizing a single \textit{scalar} reward \cite{silver2021reward}. A growing body of work contests the sufficiency of this framing, arguing that many objectives of interest cannot always be expressed by a scalar reward \cite{vamplew2022scalar, skalse2023the, bowling2023settling, van2017hybrid} and are more naturally represented as \textit{vectors} \cite{vamplew2022scalar, skalse2023the}. Vector rewards and Pareto-optimal policy sets are common in multi-objective decision-making \cite{roijers2013survey,hayes_practical_2022}, where general policy optimization recovers the Pareto front by conditioning a single policy-network on a sampled weight vector. This literature, however, largely assumes the ground-truth vector reward function is given. The reward-free viewpoint of \cite{chen2026a} relaxes part of this assumption by using preference-guided exploration as an auxiliary task, yet still requires the ground-truth multi-objective reward during training. In contrast, LEMUR avoids these assumptions. Instead, it posits that true objectives are never directly observed and must be learned, in this case from preference feedback. Thus, the agent jointly learns reward models and an optimized policy that balances its various components.

\paragraph{Interactive, Non-Linear, and Distributional MORL}
Several lines of MORL work go beyond recovering a linear coverage set, but all assume the objective rewards are given. Interactive MORL elicits a user's preferences to select among trade-offs: Promptable Behaviors \cite{hwang_promptable_2023} trains a single weight-conditioned policy on known sub-rewards and infers a user's weights from demonstrations, trajectory comparisons, or language, and human-in-the-loop preference-based MORL \cite{hu2026humanintheloop} steers policy search toward a decision-maker's preferred region of the front through preference queries. Non-linear MORL optimizes concave or otherwise non-linear utilities of the return, distinguishing the scalarized expected return from the expected scalarized return \cite{ruadulescu2020multi,agarwal_multi-objective_2022,3709347.3743798}, and distributional MORL recovers Pareto-optimal return distributions rather than expectations \cite{cai2023distributional}. These directions are complementary to LEMUR: our non-linear scalarization experiment (Appendix~\ref{app:scalarisation}) shows that LEMUR's learned per-teacher rewards can be optimized under a non-linear welfare function, and pursuing expected-scalarized or distributional objectives over \textit{learned} rewards is a promising direction for future work.

\paragraph{Multi-Dimensional and Fine-Grained Preference Learning.}
Standard RLHF distills human comparisons into a single scalar reward \cite{christiano_deep_2017,ouyang2022training}, which can conflate distinct criteria and collapse the multi-objective structure of human values \cite{pmlr-v235-sorensen24a}. A line of work therefore decomposes feedback into finer-grained components: Fine-Grained RLHF \cite{wu2023fine} attaches rewards to localized segments and categories, while Rewarded Soups \cite{rame2023rewarded} and MODPO \cite{zhou2024beyond} obtain a separate signal per objective and expose a Pareto family through weight interpolation or scalarized preference optimization. These methods combine per-objective signals \textit{post hoc} rather than jointly learning the rewards and a trade-off policy, and are developed almost exclusively for LLM post-training. A related direction forgoes the reward model entirely, optimizing the policy directly from preference data (e.g., DPO \cite{rafailov2023direct}); such methods are effective but train on static or periodically refreshed preferences, forgoing the online, interactive feedback and policy optimization, ideal for reliable reward learning under policy improvement \cite{christiano_deep_2017, pmlr-v139-lee21i, gao2023scaling,bai2022training, ross2011reduction, ibarz2018reward, stiennon2020learning}. 

A further perspective treats heterogeneous feedback as hidden context, learning a distribution over reward functions \cite{siththaranjan2024distributional}; this captures \textit{which} preferences are present but does not, on its own, optimize an explicit compromise between them. LEMUR differs from all three: it learns an explicit, objective-specific reward model per feedback stream \textit{online}, and jointly optimizes a set of policies over the resulting vector reward in continuous control.
A complementary line moves beyond the Bradley-Terry preference model assumption itself. General preference models represent preferences with a richer (e.g., skew-symmetric, $k$-dimensional) structure that admits \textit{intransitive} cycles, optimizing policies directly against this structure rather than a scalar reward \cite{umer2026general}. Related game-theoretic formulations cast alignment as a Nash or Stackelberg equilibrium over preferences \cite{pmlr-v235-munos24a, pasztor2025stackelberg}. Intransitivity is a different failure mode of scalar rewards from the one we study; it concerns the \textit{shape} of preferences rather than the presence of multiple objectives, and we regard bridging it with multi-objective preference optimization as promising future work. In the LLM setting, rubric-based rewards have similarly been used to supply multi-dimensional supervision for reinforcement fine-tuning, though their reliability is sensitive to rubric coverage and correlated criteria \cite{gunjal2025rubrics, shen2026rethinking}. These directions are largely orthogonal to LEMUR, which targets online vector-reward learning and trade-off policy optimization.
\paragraph{Connections to Diversity and Foundation-Model Post-Training.}
A contemporary line of work observes that scalar RL post-training induces \textit{entropy collapse}, eroding the solution diversity required by inference-time search \cite{kirk2024understanding}. Most directly, Vector Policy Optimization \cite{bahlous2026vector} samples scalarizations over the reward simplex and trains a policy to output a set of solutions spanning the Pareto front, improving downstream best-of-$N$ search. Such methods share LEMUR's premise that collapsing a vector reward into a scalar discards useful structure. They differ from our setting in three respects: (i) they target LLM \textit{test-time search}, seeking a diverse candidate pool for a downstream selector, whereas LEMUR seeks a policy that \textit{compromises} between objectives \textit{during deployment}; (ii) they assume the reward components are \textit{known and observable} (e.g., per-test-case correctness), whereas LEMUR must \textit{learn} them from feedback; and (iii) they operate on natural language generation rather than continuous control. We therefore treat this literature as motivating context rather than comparable works, and leave the study of train-time multi-objective learning from feedback to improve test-time diversity \cite{ding2024quality,pierrot2022multi} as future work.
\paragraph{Pluralistic Alignment and Multiple Principals.}
LEMUR's multi-teacher formulation connects to \textit{pluralistic alignment}, which holds that a single reward model cannot represent the plurality of human values \cite{pmlr-v235-sorensen24a} and that an agent should instead balance multiple objectives to reach a compromise across them. A closely related framing casts this as serving \textit{multiple principals}: a single agent acting on behalf of several stakeholders must respect the preferences of each, turning alignment into the problem of representing and trading off competing principals rather than satisfying one \cite{fickinger2020multi}. How to reconcile these principals is itself contested. Naively aggregating preferences via RLHF has been shown to behave as a Borda count over latent objectives, with limited normative justification \cite{siththaranjan2024distributional}, and others question whether aggregation is the right frame at all, proposing non-aggregative alternatives for reconciling plural values \cite{zhi2025beyond}. Practical efforts that do aggregate rely on curated consensus datasets or hand-specified aggregation functions \cite{bakker2022fine,rodriguez2023multi,tessler_ai_2024}, both costly to obtain a priori and brittle when preferences shift during deployment. Rather than collapsing principals to a consensus in advance or assuming their reward functions are known, LEMUR preserves each principal's objective as a separate learned reward model and recovers an explicit compromise through multi-objective optimization.


\section{Additional Experiments}
\label{app:additionalexperiments}


\subsection{Ablation: Query Length}
\label{app:query_length}

\begin{figure}[tbp]
    \centering
    \includegraphics[width=0.72\linewidth]{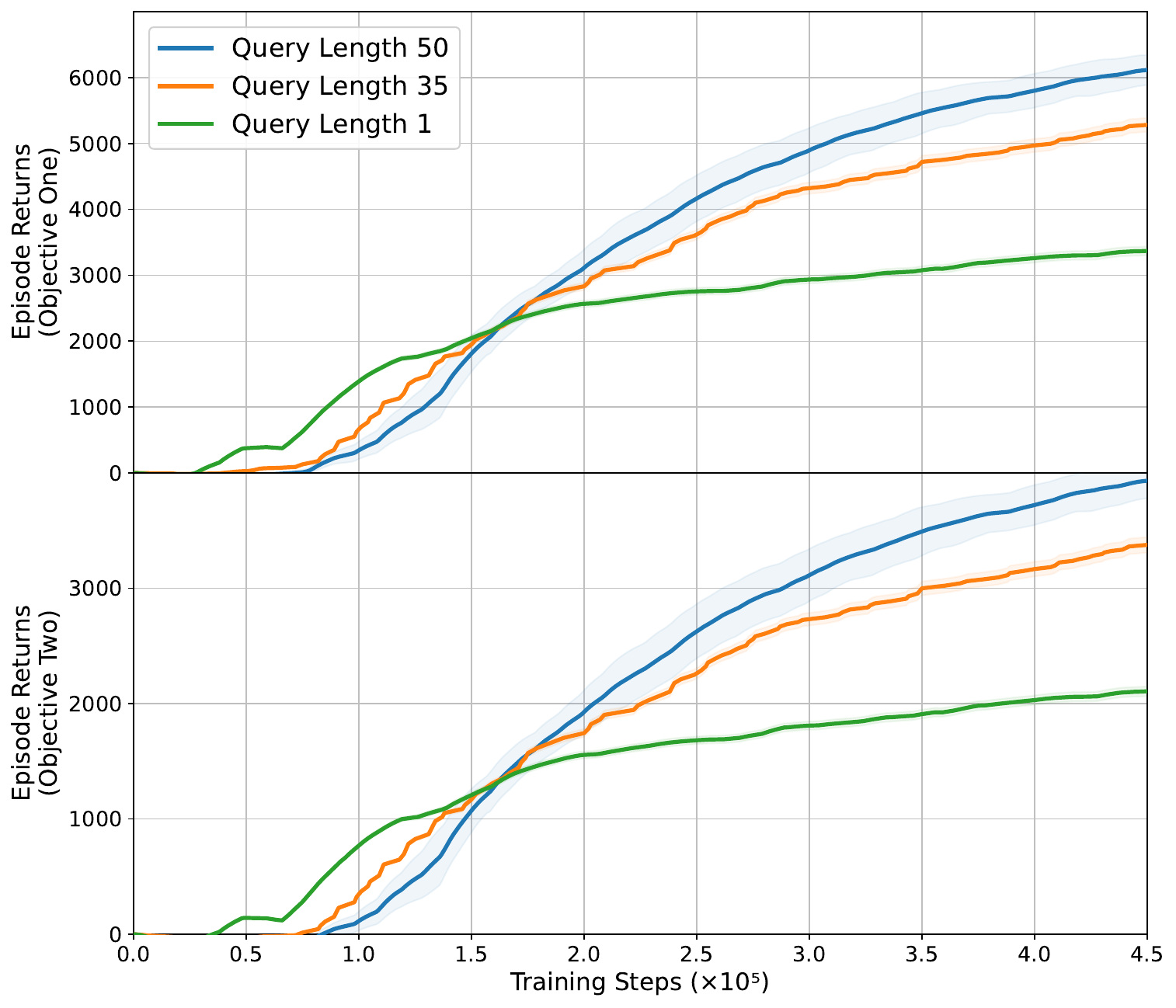}
    \caption{\textbf{Effect of query segment length (\texttt{MO-Cheetah}).} Ground-truth return
    per objective for preference queries of length $50$ (default), $35$ and $1$ transition, at a
    fixed budget of $300$ queries per teacher. Shading is $\pm1$ std. Shorter segments learn faster
    early but plateau by ${\sim}2 \times 10^5$ steps, while length $50$ overtakes them and
    continues improving. Mean $\pm$ std over five random seeds.}
    \label{fig:query_length_ablation}
\end{figure}

LEMUR elicits preferences over \textit{trajectory segments}, determining the behavioral context teachers see per comparison. At a fixed budget of 300 queries per teacher, we tested segment lengths of 50, 35, and 1 transition (Figure~\ref{fig:query_length_ablation}). LEMUR maintains performance even with shorter segments of feedback. The improvement in performance due to longer segments is potentially due to more context for reward learning.

\subsection{Ablation: Weight-Conditioned Reward Model vs.\ Reward Ensemble}
\label{app:wcvsensemble}

\begin{figure}[htbp]
    \centering
    \includegraphics[width=0.85\linewidth]{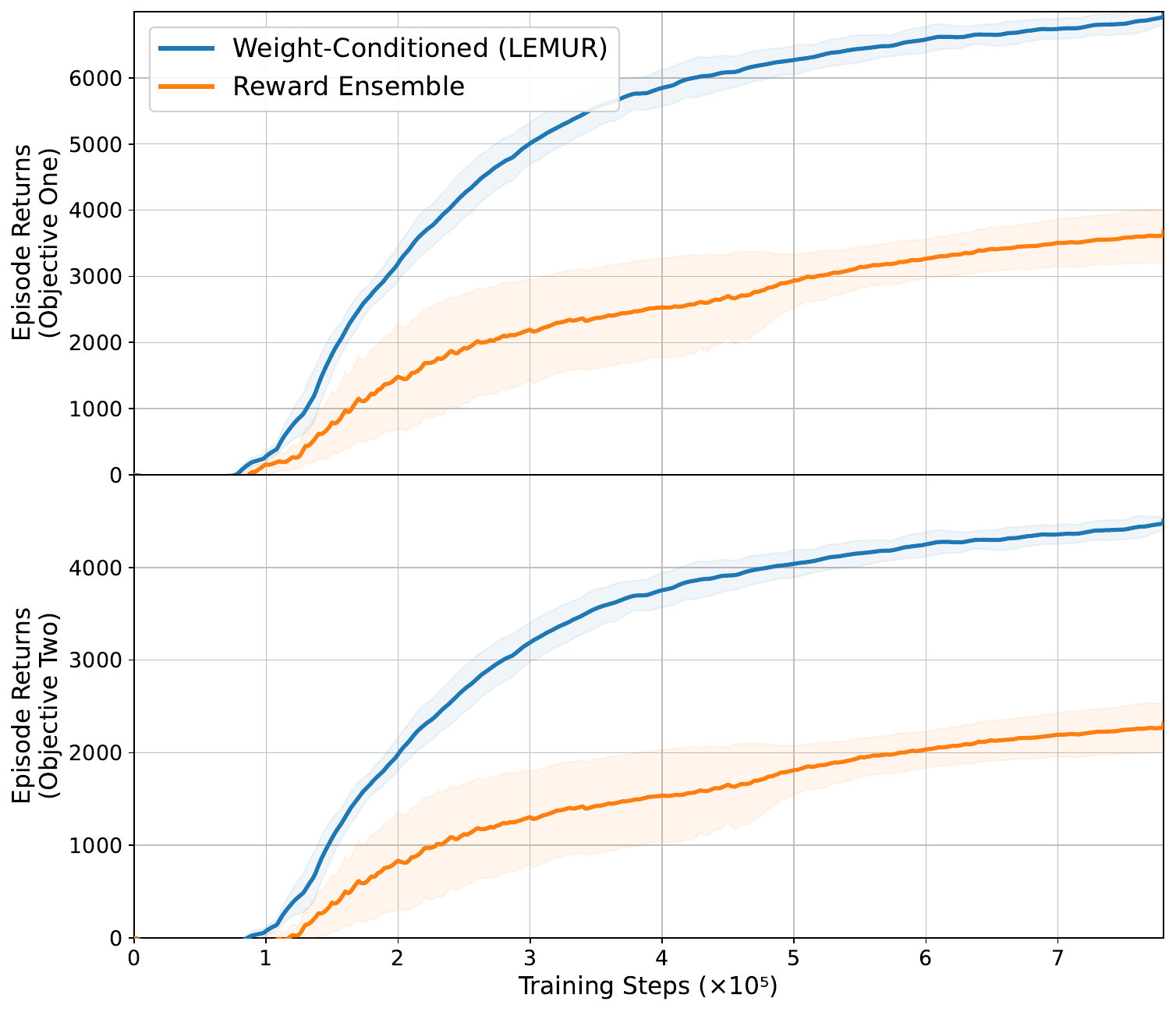}
    \caption{\textbf{Reward-model ablation (\texttt{MO-Cheetah}).} LEMUR with the weight-conditioned reward model versus an ensemble of unconditioned reward models, under an identical optimizer, explorer, and query budget. Mean $\pm$ std over five seeds.}
    \label{fig:wc_vs_ensemble}
\end{figure}
We replace each teacher's weight-conditioned model $\hat{\mathbf{R}}_{\psi_j}$ with an ensemble of three unconditioned MLPs, each outputting a scalar reward and trained by standard Bradley-Terry cross-entropy on that teacher's labels, as in single-objective PbRL \cite{christiano_deep_2017,lee2021bpref,pmlr-v139-lee21i}. Everything else is held fixed: the MORL/D backbone, explorer, query budget, and segment length. The learner in this variant never uses the teacher profiles $\boldsymbol{\omega}_j$, neither in the preference likelihood nor in relabeling.

Figure~\ref{fig:wc_vs_ensemble} shows two results. First, without the profiles, LEMUR still learns a compromise policy that improves steadily on both objectives, so the per-teacher decomposition and online relabeling are sufficient to resolve conflicting feedback. Second, weight conditioning yields a gain with smaller variance across multiple runs.

We refer to \citet{mu2025preference} and \citet{yang2019generalized} for further discussion of weight-conditioned multi-objective reward and value learning.

\subsection{Ablation: Query Sampling Strategy}
\label{app:query_sampling}

\begin{figure}[tbp]
    \centering
    \includegraphics[width=0.72\linewidth]{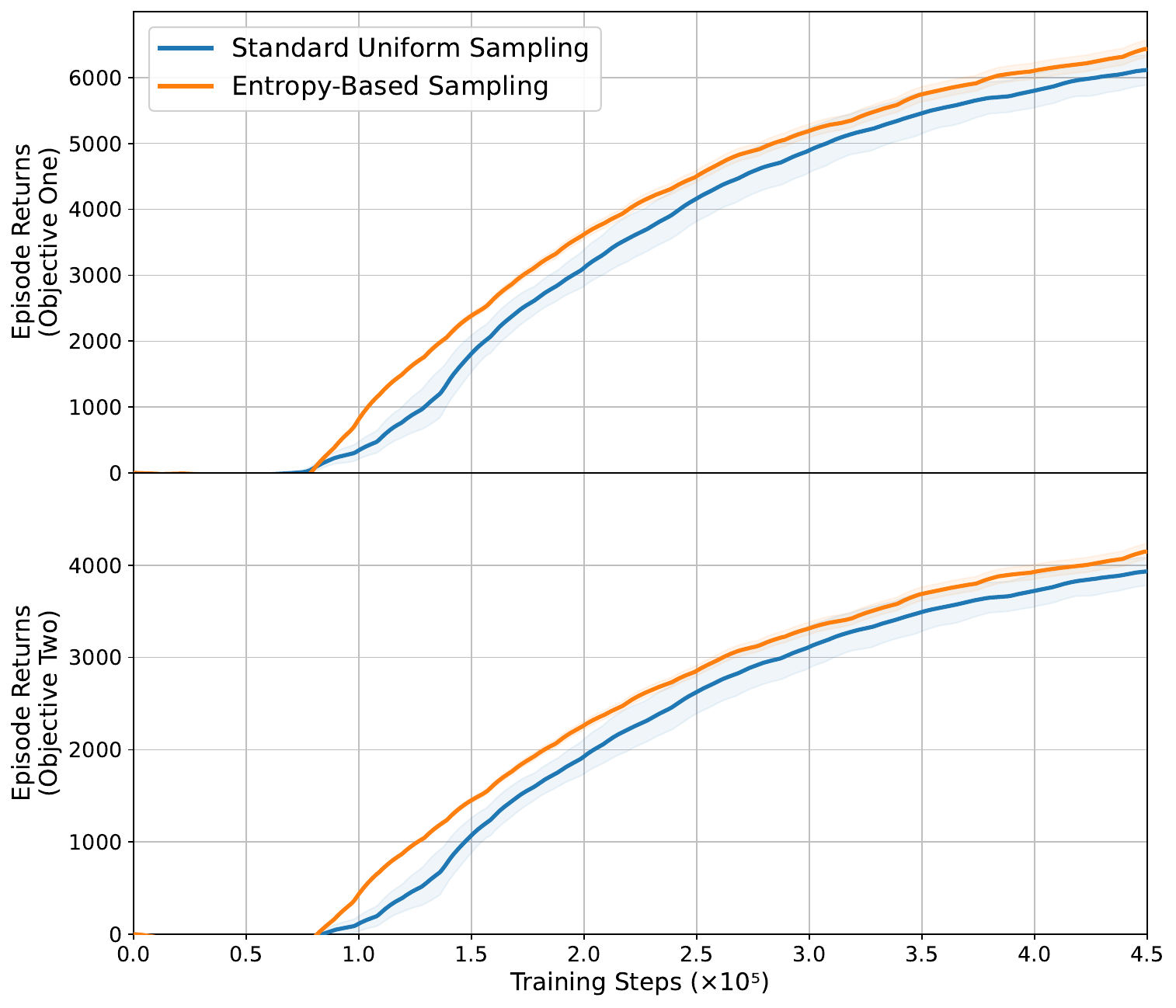}
    \caption{\textbf{Uniform vs.\ entropy-based query sampling (\texttt{MO-Cheetah}).}
    Ground-truth return per objective for the default uniform sampler and an entropy-based
    sampler that scores a $10\times$ candidate pool by reward-model uncertainty.  Entropy-based
    selection is consistently ahead, but by a small margin relative to the effect of query
    length (Figure~\ref{fig:query_length_ablation}). Mean $\pm$ std over five random seeds.}
    \label{fig:query_sampling_ablation}
\end{figure}

We also tested whether \textit{which} segments are queried matters at a fixed budget and length. We compared LEMUR's default uniform sampler against an entropy-based sampler that selects the most uncertain segment pairs from a 10$\times$ candidate pool (Figure~\ref{fig:query_sampling_ablation}).

Entropy-based selection yields a modest gain in performance. Uniform sampling recovers most of the achievable return without requiring uncertainty estimates, candidate pools, or extra forward passes. Combined with our query-length findings, this shows that under a fixed budget, \textit{how much behavior each query covers} matters significantly more than \textit{which} specific segments are chosen.

\begin{figure}[tbp]
    \centering
    \includegraphics[width=0.85\linewidth]{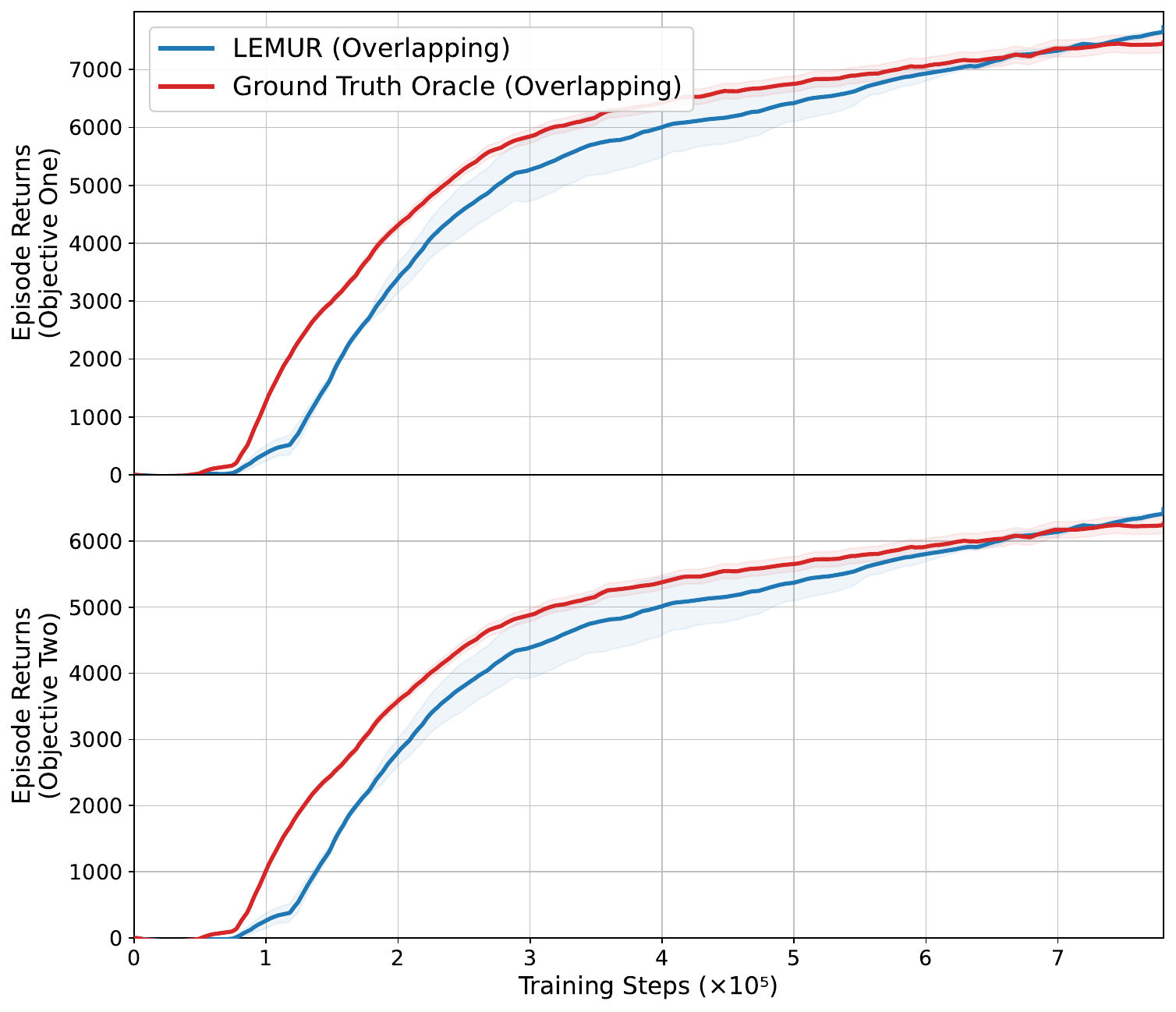}
    \caption{\textbf{Overlapping preferences (\texttt{MO-Cheetah}).} LEMUR converges close to the ground-truth oracle on both objectives. Mean $\pm$ std over five seeds.}
    \label{fig:overlapping}
\end{figure}
\subsection{Overlapping Preferences}
\label{app:overlapping}

Our main results consider teachers with conflicting preferences. A natural question is
whether the method required to resolve conflict imposes a cost when the teachers largely
\emph{agree}. We therefore repeat the experiment with overlapping (near-aligned) teacher weight
vectors, comparing LEMUR against the ground-truth oracle upper bound under the same overlap
condition.

Figure~\ref{fig:overlapping} shows that LEMUR tracks the oracle closely on both objectives,
converging to a comparable final return with only a modest lag in sample efficiency. This
indicates that the weight-conditioned formulation degrades gracefully toward the
single-preference case, when there is little conflict to resolve.

\subsection{Robustness to the Level of Teacher Conflict}
\label{app:conflict_levels}
We evaluate whether LEMUR remains effective as the teachers' preferences diverge further. On \texttt{MO-Cheetah}, we increase the conflict between the two teachers from the standard setting used in our main experiments ($\boldsymbol{\omega}_A = [0.6,0.4]$, $\boldsymbol{\omega}_B = [0.4,0.6]$) to Hard ($[0.75,0.25]$, $[0.25,0.75]$) and Harder ($[0.9,0.1]$, $[0.1,0.9]$), and compare LEMUR against the Oracle trained on ground-truth rewards at each level. Since each level changes the teachers' utilities, and hence the scale of their returns, we report normalised returns, $\tilde{R} = (R - R_{\text{random}})/(R_{\text{Oracle}} - R_{\text{random}})$, computed per teacher and level, so that 0 corresponds to a uniform-random policy and 1 to the Oracle.

\begin{figure}[htbp]
    \centering
    \includegraphics[width=0.85\linewidth]{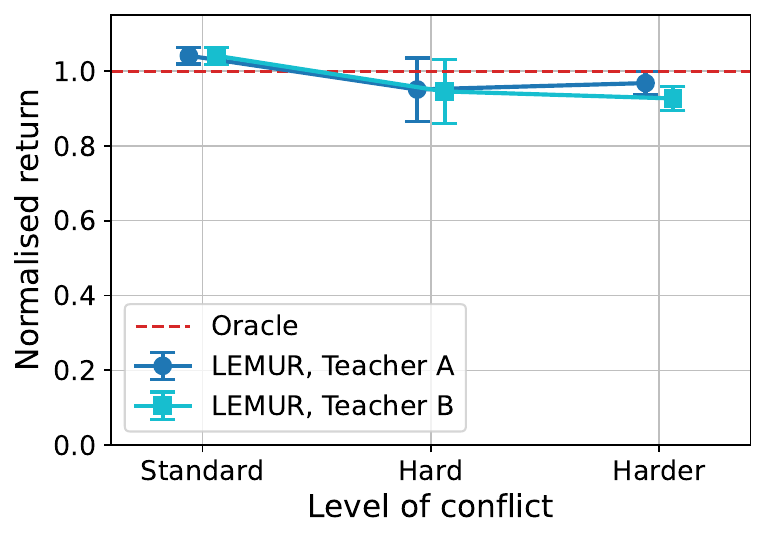}
    \caption{LEMUR under increasing teacher conflict on \texttt{MO-Cheetah}: Standard ($\boldsymbol{\omega}_A = [0.6,0.4]$, $\boldsymbol{\omega}_B = [0.4,0.6]$), Hard ($[0.75,0.25]$, $[0.25,0.75]$) and Harder ($[0.9,0.1]$, $[0.1,0.9]$). Normalized returns per teacher (0: uniform-random policy, 1: Oracle); mean $\pm$ standard deviation over five seeds.}
    \label{fig:conflict_levels}
\end{figure}

Figure~\ref{fig:conflict_levels} shows that LEMUR remains close to the Oracle for both teachers as conflict increases, retaining over $0.9$ of the Oracle's return at every level. This indicates that learning a separate reward model per teacher remains effective as their preferences diverge, without any change to the method or its hyperparameters.

\begin{figure}[t]
  \centering
  \includegraphics[width=0.85\linewidth]{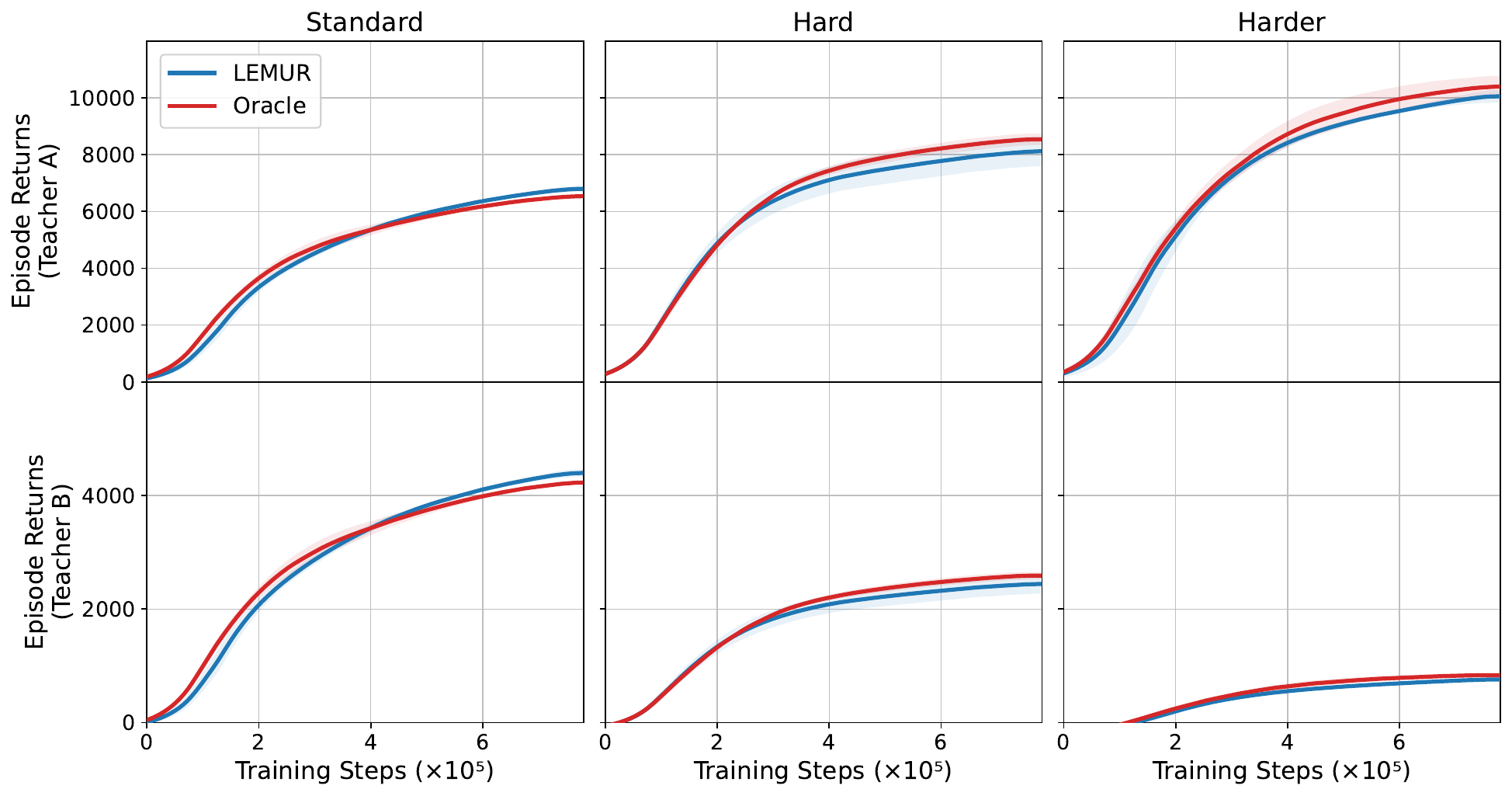}
  \caption{LEMUR and the Oracle on MO-HalfCheetah at increasing teacher conflict. Teacher weights: Standard $A{=}[0.6,0.4]$, $B{=}[0.4,0.6]$; Hard $A{=}[0.75,0.25]$, $B{=}[0.25,0.75]$; Harder $A{=}[0.9,0.1]$, $B{=}[0.1,0.9]$. Mean $\pm$ standard deviation over five seeds.}
  \label{fig:conflict_curves_halfcheetah}
\end{figure}

\begin{figure}[t]
  \centering
  \includegraphics[width=0.85\linewidth]{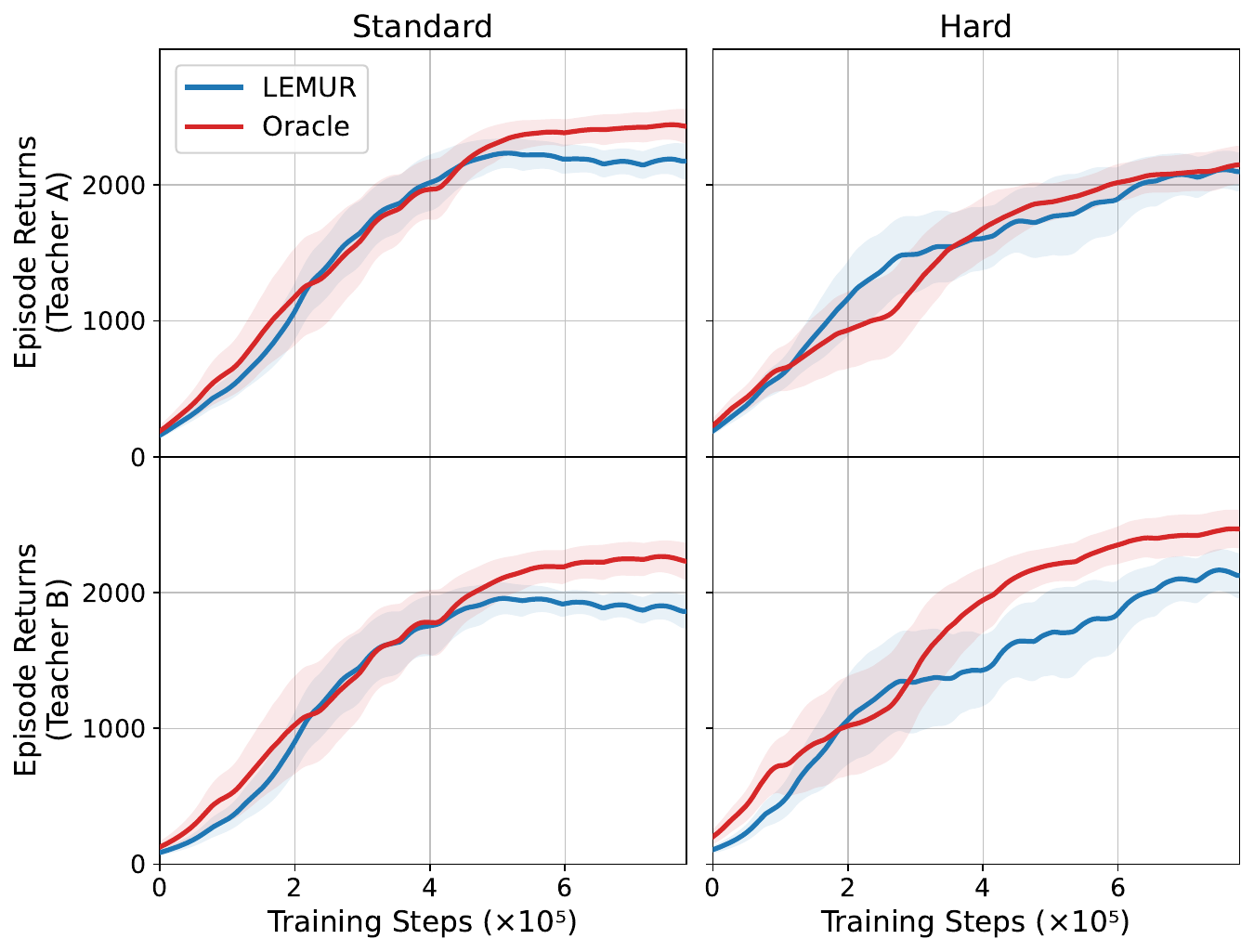}
  \caption{LEMUR and the Oracle on MO-Hopper at increasing teacher conflict. Teacher $A{=}[0.8,0.1,0.1]$ at both levels; teacher $B{=}[0.3,0.5,0.2]$ (Standard) and $[0.1,0.8,0.1]$ (Hard). Mean $\pm$ standard deviation over five seeds.}
  \label{fig:conflict_curves_hopper}
\end{figure}
\paragraph{Increasing teacher conflict.}
We increase the conflict between the two teachers by moving their weight vectors further apart, and compare LEMUR with the Oracle, which is trained on the teachers' ground-truth utilities. On MO-HalfCheetah (Figure~\ref{fig:conflict_curves_halfcheetah}), LEMUR matches or slightly exceeds the Oracle at the standard level and remains within 10\% of it for both teachers at the hard and harder levels. On MO-Hopper (Figure~\ref{fig:conflict_curves_hopper}), LEMUR reaches 81--87\% of the Oracle's return at the standard level and 86--97\% at the hard level, so the gap to the Oracle does not widen as conflict increases. Returns are not comparable across levels, as each level changes the teachers' utilities.

\subsection{Weight Adaptation and Weight Sampling}
\label{app:weight_sampling}
LEMUR initializes its population with equally spaced weight vectors and adapts them online with Pareto Simulated Annealing (PSA) \cite{felten_multi-objective_2024}. To test how sensitive LEMUR is to this choice, we compare against two alternatives on \texttt{MO-Cheetah}: (i) \textit{No PSA}, where each policy keeps its initial weight vector fixed throughout training; and (ii) \textit{Per-Update Weight Sampling}, where every gradient update scalarises the actor and critic losses with a weight vector sampled uniformly from the simplex, $\mathbf{w} \sim \mathrm{Dir}(\mathbf{1})$, rather than with the policy's own weight. All other components of LEMUR are unchanged.

\begin{figure}[htbp]
    \centering
    \includegraphics[width=0.85\linewidth]{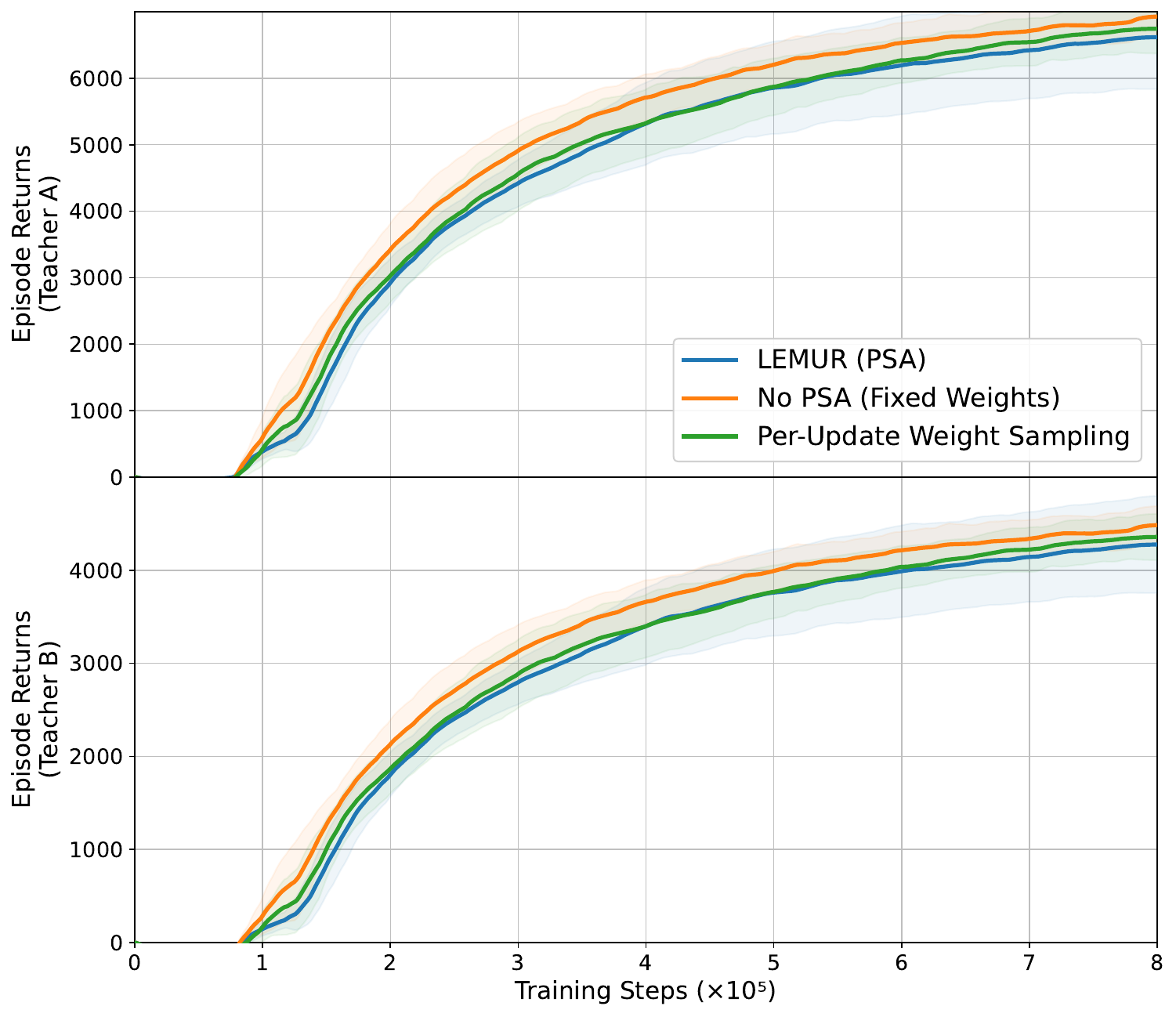}
    \caption{LEMUR with different weight adaptation schemes on \texttt{MO-Cheetah}: PSA weight adaptation (default), fixed weights without PSA, and per-update weight sampling from $\mathrm{Dir}(\mathbf{1})$. Ground-truth returns for each teacher; mean $\pm$ standard deviation over five seeds.}
    \label{fig:weight_sampling}
\end{figure}

Figure~\ref{fig:weight_sampling} shows that all three variants converge to comparable returns for both teachers, with overlapping variance across seeds. LEMUR's performance is therefore robust to how scalarization weights are adapted or sampled. Even though in this environment, PSA does not seem to yield any benefit, we still maintain the use of PSA in LEMUR, due to stability and improved optimization coverage in other environments.

\subsection{Robustness to Label Noise: LEMUR vs.\ PbMORL}
\label{app:noisy_pbmorl}
\begin{figure}[htbp]
    \centering
    \includegraphics[width=\columnwidth]{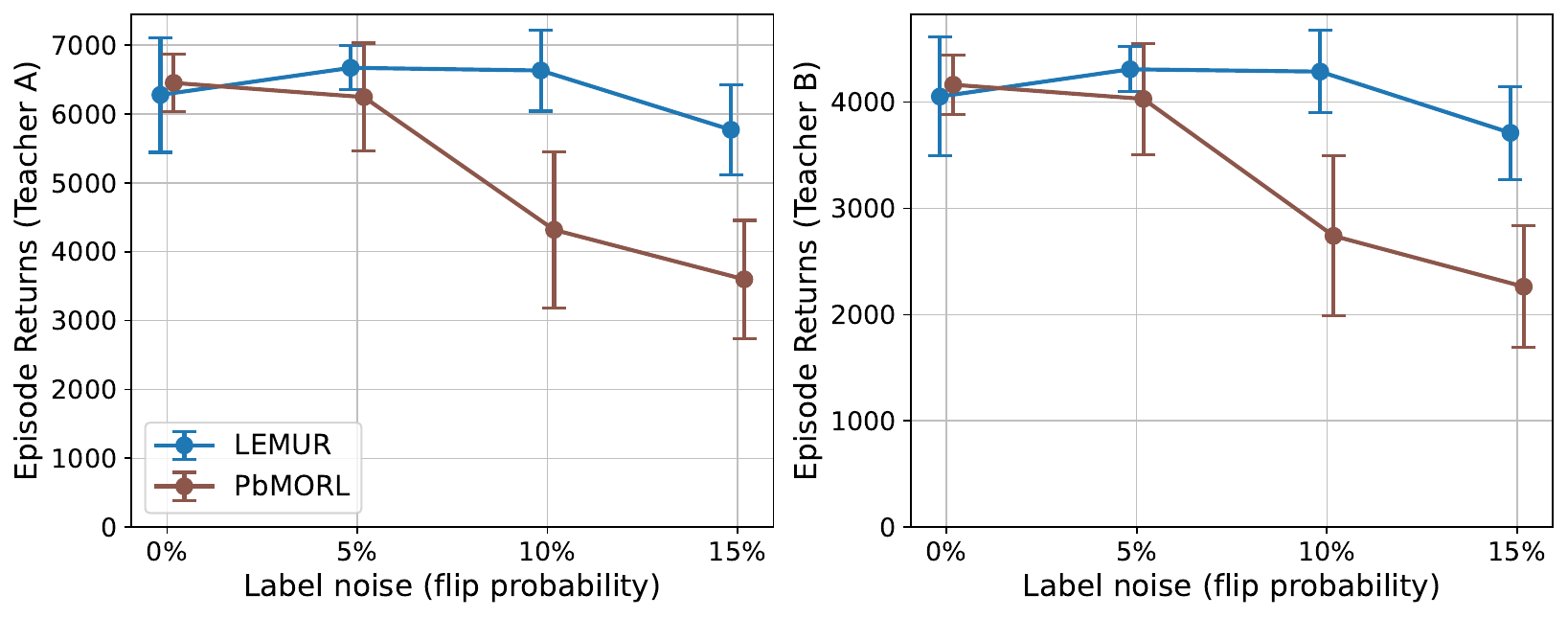}
    \caption{Converged returns of LEMUR and PbMORL on \texttt{MO-Cheetah} under increasing label noise. Mean $\pm$ standard deviation over five seeds.}
    \label{fig:noisy_pbmorl}
\end{figure}

We repeat the label-noise study of Figure~\ref{fig:three_ablations}(b) for PbMORL, the strongest baseline in this environment, flipping each teacher's preference labels with probability up to 15\% \cite{lee2021bpref}. Figure~\ref{fig:noisy_pbmorl} shows that LEMUR's performance remains relatively robust with increasing noise, whereas PbMORL degrades. It should be noted that we kept the strongest hyperparameters for both these methods.  .

\subsection{Non-Linear Scalarization}
\label{app:scalarisation}
LEMUR's policy optimization uses linear scalarization of the learned reward vector (Equation~\ref{lemurobjective}). To test whether the framework extends beyond it, we replace the weighted sum with the Generalized Gini Welfare (GGF) \cite{weymark1981generalized,siddique2023fairness}, a concave, non-linear scalarization that favors the worst-off teacher: $f_{\boldsymbol{\alpha}}(\hat{\mathbf{r}}) = \sum_{i=1}^{m} \alpha_i \hat{r}_{(i)}$, where $\hat{r}_{(1)} \leq \dots \leq \hat{r}_{(m)}$ are the per-teacher learned rewards sorted in increasing order and $\alpha_i \propto 2^{-(i-1)}$ are strictly decreasing. Under GGF, every policy in the population optimizes this single welfare objective, targeting the most equitable compromise rather than a spread of trade-offs. All other components of LEMUR are unchanged.

\begin{figure}[htbp]
    \centering
    \includegraphics[width=0.85\linewidth]{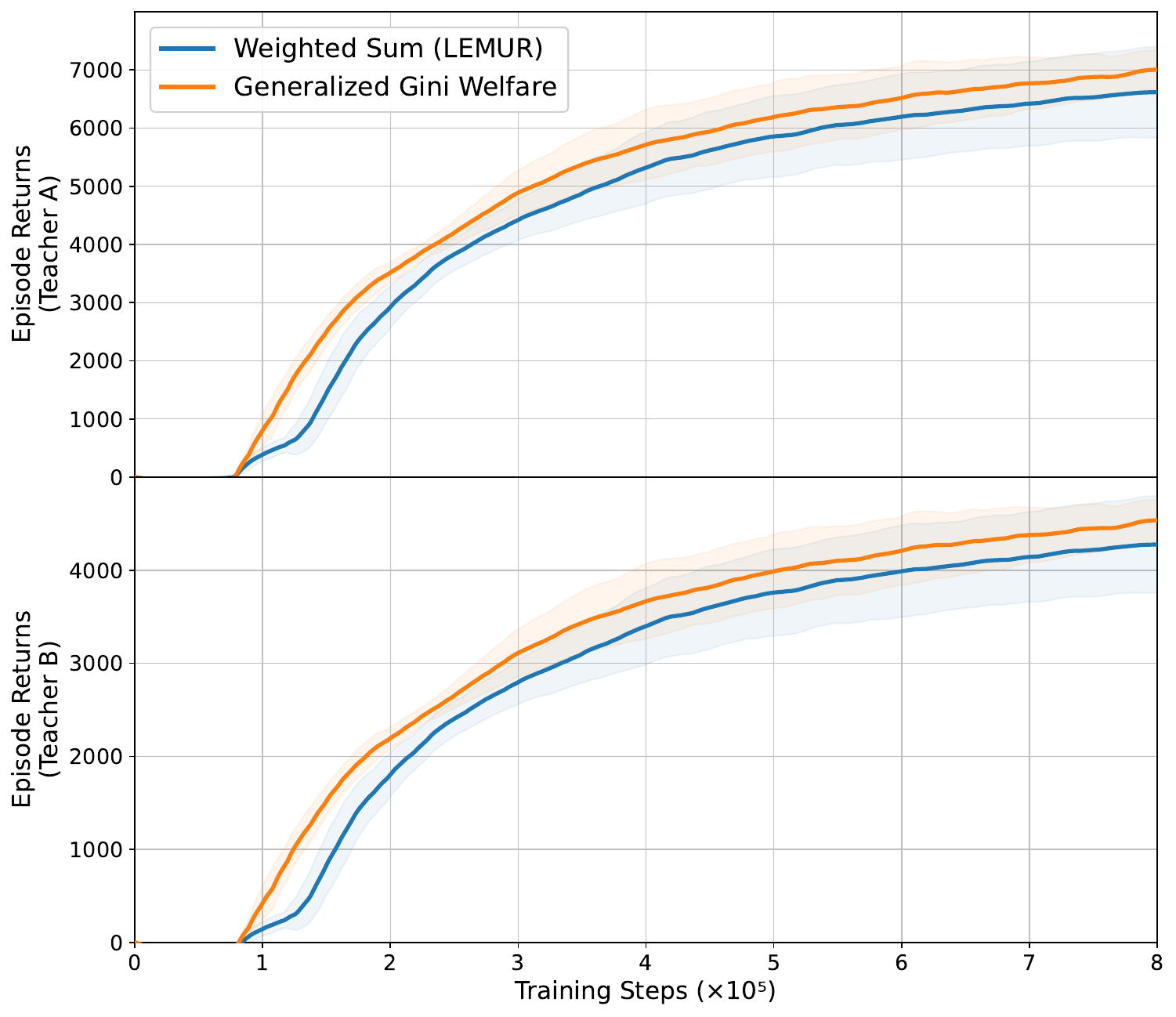}
    \caption{LEMUR with linear (weighted sum, as in the main experiments) and non-linear (Generalized Gini Welfare) scalarisation on \texttt{MO-Cheetah}. Ground-truth returns for each teacher; mean $\pm$ standard deviation over five seeds.}
    \label{fig:scalarisation}
\end{figure}

Figure~\ref{fig:scalarisation} shows that LEMUR learns effectively under both scalarizations on \texttt{MO-Cheetah}, with GGF performing comparably. This indicates that LEMUR's learned per-teacher reward models are compatible with non-linear, fairness-oriented scalarizations, addressing the extension noted in our limitations. 

\subsection{Importance of Off-Policy Optimization}
\label{app:lemur_ppo}
\begin{figure}[htbp]
    \centering
    \includegraphics[width=0.85\linewidth]{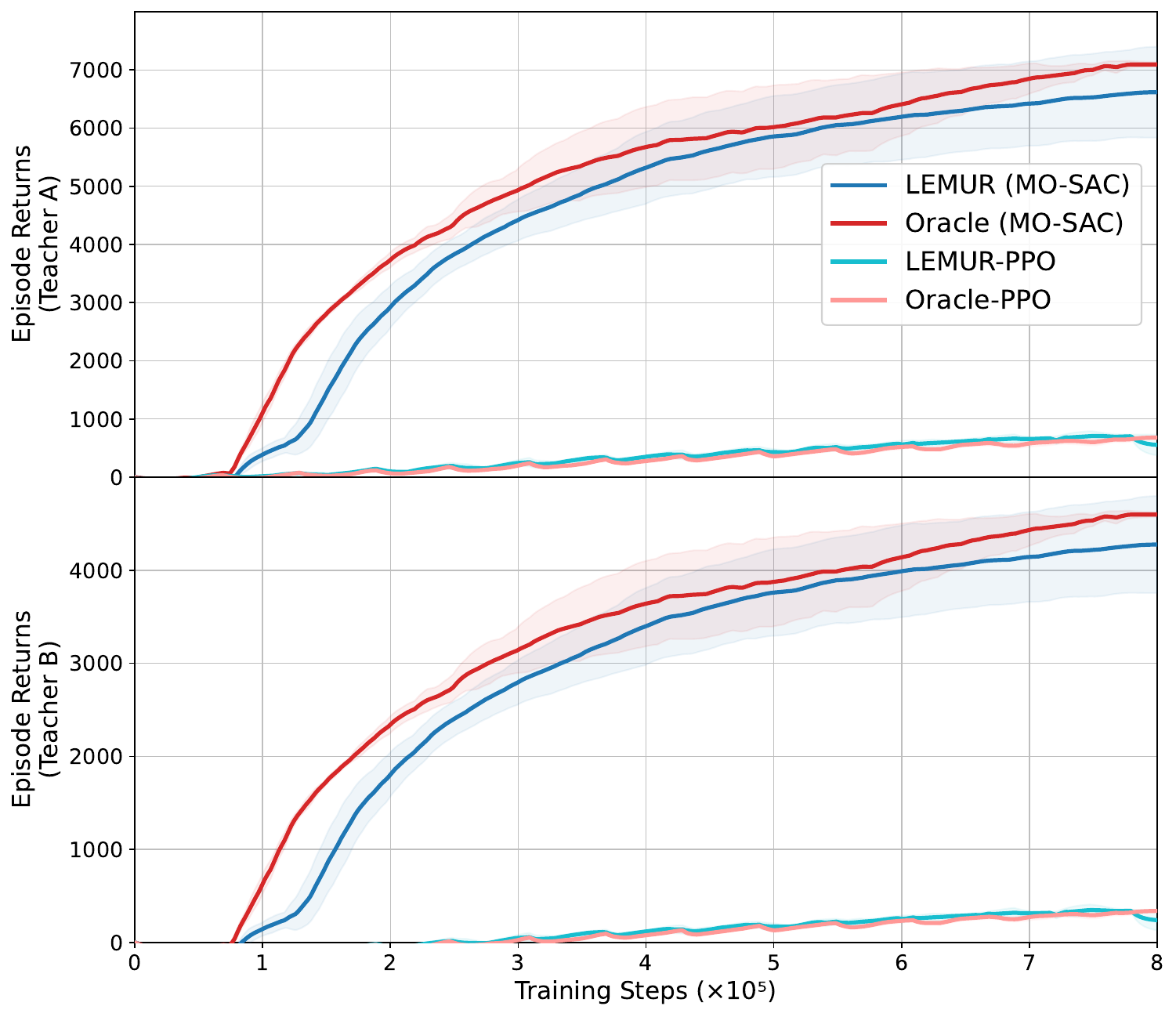}
    \caption{LEMUR and the Oracle on \texttt{MO-Cheetah} with off-policy (MO-SAC) and on-policy (PPO) policy optimisation. Mean $\pm$ standard deviation over five seeds}
    \label{fig:lemur_ppo}
\end{figure}

LEMUR uses an off-policy learner (MO-SAC) to reuse experience under a limited feedback budget; the importance of off-policy learning was shown in prior single-objective PbRL work \cite{pmlr-v139-lee21i}. To test this choice, we replace it with PPO \cite{schulman_proximal_2017} within the same MORL/D framework (LEMUR-PPO), using standard MuJoCo PPO settings, and compare against an Oracle trained with PPO on ground-truth rewards. Figure~\ref{fig:lemur_ppo} shows that LEMUR-PPO and Oracle-PPO learn far more slowly than their off-policy counterparts within the same environment-step budget, supporting our choice of an off-policy optimizer for sample efficiency.

\subsection{Post-hoc Weight Interpolation of Single-Objective Policies}
\label{app:soups}
A potential alternative to learning a set of compromise policies is to simply \textit{interpolate} between the single-objective policies of each teacher. Combining models post hoc by interpolating their weights has been investigated, notably for LLM alignment \cite{rame2023rewarded,rame_warm_2024}. We conduct a simple experiment to test whether this approach yields suitable compromise policies. On \texttt{MO-Cheetah}, with the scripted teachers of our main experiments ($\boldsymbol{\omega}_A = [0.6, 0.4]$, $\boldsymbol{\omega}_B = [0.4, 0.6]$), we train a single-objective SAC policy on each teacher's learned reward model, with both policies of a pair sharing the same initialization. For $\lambda \in [0,1]$ we interpolate all actor parameters, $\theta_\lambda = (1-\lambda)\,\theta_A + \lambda\,\theta_B$, and evaluate each interpolated policy on both teachers' ground-truth utilities, $u_j = \sum_t \boldsymbol{\omega}_j^\top \mathbf{r}(s_t, a_t)$.

Table~\ref{tab:soups} shows that interpolation does not yield adequate compromise returns as evaluated on the ground-truth scripted teacher return. Every interpolated policy, namely the midpoint policy ($\lambda = 0.5$), recovers poor returns. It seems that Weight averaging relies on fine-tuned models remaining in a shared, linearly connected region of parameter space \cite{rame2023rewarded}.

\begin{table}[htbp]
  \centering
  \small
  \begin{tabular}{lcc}
    \toprule
    \textbf{Policy} & \textbf{Teacher A return} & \textbf{Teacher B return} \\
    \midrule
    $\lambda = 0.2$ & $2332.0 \pm 548.9$ & $1485.3 \pm 359.7$ \\
    $\lambda = 0.4$ & $1029.0 \pm 41.7$ & $635.9 \pm 27.5$ \\
    $\lambda = 0.5$ & $819.6 \pm 65.0$ & $497.8 \pm 44.1$ \\
    $\lambda = 0.6$ & $896.4 \pm 180.3$ & $546.2 \pm 119.1$ \\
    $\lambda = 0.8$ & $2428.1 \pm 299.1$ & $1547.7 \pm 195.3$ \\
    \bottomrule
  \end{tabular}
  \caption{Weight interpolation of single-objective SAC policies trained on each scripted teacher's learned reward, $\theta_\lambda = (1-\lambda)\theta_A + \lambda\theta_B$, on \texttt{MO-Cheetah}. Mean $\pm$ standard deviation of each teacher's ground-truth episodic return over five seeds and 10 evaluation episodes.}
  \label{tab:soups}
\end{table}

\subsection{Welfare of the Learned Compromise}
\label{app:welfare}
To quantify how well the learned policies serve the teachers jointly, we report the Generalized Gini Welfare (GGF) \cite{weymark1981generalized}, as used to evaluate fairness in preference-based RL \cite{siddique2023fairness}. For a policy rollout $\tau$, let $u_j(\tau) = \sum_{t} \boldsymbol{\omega}_j^\top \mathbf{r}(s_t, a_t)$ denote teacher $j$'s ground-truth utility, and let $u_{(1)} \leq \dots \leq u_{(m)}$ be these utilities sorted in increasing order. The GGF is
\begin{equation}
\label{eq:ggf}
\phi_{\boldsymbol{\alpha}}(\mathbf{u}) = \sum_{i=1}^{m} \alpha_i \, u_{(i)}, \qquad \alpha_i = \frac{2^{-(i-1)}}{\sum_{k=1}^{m} 2^{-(k-1)}},
\end{equation}
where the strictly decreasing weights $\alpha_1 > \dots > \alpha_m$ place the largest weight on the worst-off teacher, so that, among utility vectors with equal total, a more balanced vector attains a higher welfare. Since both LEMUR and PbMORL learn a set of policies, we evaluate each policy periodically on the ground-truth rewards and report the GGF of the best compromise in the set, i.e., the rollout maximizing Equation~\ref{eq:ggf}. Table~\ref{tab:welfare} reports the converged welfare, averaged over the final 10\% of training, on \texttt{MO-Cheetah}. LEMUR attains a substantially higher welfare than PbMORL with lower variance across seeds, indicating that its learned set contains a compromise policy that serves both teachers better simultaneously.

\begin{table}[htbp]
  \centering
  \small
  \begin{tabular}{lc}
    \toprule
    \textbf{Method} & \textbf{GGF Welfare ($\uparrow$)} \\
    \midrule
    \textbf{LEMUR}  & $\mathbf{4886 \pm 412}$ \\
    PbMORL          & $3313 \pm 929^{}$ \\
    \bottomrule
  \end{tabular}
  \caption{Converged Generalized Gini Welfare (Equation~\ref{eq:ggf}) of the best compromise policy on \texttt{MO-Cheetah}, reported as mean $\pm$ standard deviation across five seeds.}
  \label{tab:welfare}
\end{table}

We report welfare as a complementary diagnostic of the learned compromise rather than as an objective that LEMUR optimizes or a criterion that this work investigates. LEMUR is designed to recover the set of trade-offs between conflicting teachers, not to prescribe which trade-off is fair. Fairness is a criterion in its own right, and the choice of welfare function, the treatment of the worst-off teacher, and the legitimacy of aggregating diverse preferences are normative questions studied in social choice, fair multi-objective RL \cite{3545946.3598870,siddique_fairness_2023}, and pluralistic alignment \cite{pmlr-v235-sorensen24a}. We leave a dedicated study of fairness in multi-teacher preference-based MORL to future work.

\section{LEMUR Implementation Details}
\label{app:lemur_implementation}

\textbf{Training Details.} We use MORL/D \cite{felten_multi-objective_2024}, a Multi-Objective Soft Actor-Critic (MO-SAC) implementation from the \textit{MORL-Generalization} benchmark \cite{teoh2025on}. Hyperparameters are summarized in Tables~\ref{tab:hyperparameters_stages} and~\ref{tab:hyperparameters_morl}.

\paragraph{Stage 1: Unsupervised pre-training.} An intrinsic-motivation explorer bootstraps a buffer $\mathcal{B}_0$ of environment transitions, run for 50{,}000 timesteps on LunarLander, 80{,}000 on Hopper, and 100{,}000 on Cheetah and MetaWorld. This buffer is shared identically with every baseline, so no method receives more exploration data than another, and all preference queries throughout training are drawn from it.

\paragraph{Stage 2: Reward learning.} Each of the $m$ teachers trains its own weight-conditioned reward model $\hat{\mathbf{R}}_{\psi_j}$: a 3-hidden-layer MLP with 256 units per layer and LeakyReLU($0.2$) activations, taking the concatenated state-action pair and producing a linear $d$-dimensional output. The query weight is not an input to the network; it enters the Bradley-Terry likelihood (Equation~\ref{eq:wcbt}) as the scalarization applied to the model's vector output, which is what permits the $d$ components to be recovered from scalar comparison labels. For the $n$-th query to teacher $j$, the weight is drawn as $\boldsymbol{\omega}^{(n)} \sim \mathrm{Dir}(\kappa\,\boldsymbol{\omega}_j)$ with $\kappa = 30$, and the scripted teacher labels the pair under that same $\boldsymbol{\omega}^{(n)}$ (Equation~\ref{eq:scriptedteacher}). Models are trained with Adam (batch size 64) at per-environment learning rates, for 100 epochs initially and 30 epochs per online update, over segments of length $H = 50$. The $m$ models are trained independently, each on its own teacher's data.

\paragraph{Role of the concentration $\kappa$.}
\label{app:kappa}
The concentration $\kappa$ controls how widely query weights vary around teacher $j$'s profile $\boldsymbol{\omega}_j$. As $\kappa \to \infty$, every query is issued at $\boldsymbol{\omega}_j$ and the model reduces to a standard Bradley-Terry reward on teacher $j$'s utility $u_j$. A finite $\kappa$ gives each model more than a single scalar constraint: queries under different weights constrain different linear combinations of the $d$ components. This identifies the components only \emph{locally}, within the span of weights sampled around $\boldsymbol{\omega}_j$; directions far from that neighbourhood remain weakly constrained. LEMUR does not require global identification, since relabeling uses only the projection $\boldsymbol{\omega}_j^\top \hat{\mathbf{R}}_{\psi_j}$ at the teacher's own profile, and the $d$-dimensional output serves as a structured parameterization of that utility. Learning a scalar utility directly is a valid alternative, and is exactly the ensemble variant of Appendix~\ref{app:wcvsensemble}, which also succeeds; we attribute the gain from weight conditioning to the additional local constraints acting as a regularizer, without claiming this mechanism is established. A finite $\kappa$ also reflects that a profile elicited from a real stakeholder is approximate, with $\kappa$ encoding confidence in $\boldsymbol{\omega}_j$; a smaller value trains the model to remain accurate over a wider region of trade-offs around it.

\paragraph{Known weights.} Issuing each query under a sampled $\boldsymbol{\omega}^{(n)}$ assumes the learner knows the weight under which a label was given. This is the standard preference datum in prior preference-based MORL methods, where a comparison is defined relative to a weight vector \cite{mu2025preference}, and it is weaker than the assumption already implicit in any scripted-teacher protocol, namely that the simulated teacher's utility is a known function of the ground-truth reward. LEMUR can be used without this component, and keeping the per-teacher decomposition, vector relabeling, and online joint optimization unchanged. Per-query weights can be thought of as an efficiency device for identifying the $d$ components from fewer labels, rather than a requirement of the framework.

\paragraph{Stage 3: Multi-objective policy optimization.} A population of 6 MO-SAC policies is trained on the learned reward vector $\hat{\mathbf{r}}_\psi \in \mathbb{R}^{m}$, coupled by a shared replay buffer, weighted-sum scalarization, PSA weight adaptation, and weight transfer between neighbouring policies (neighbourhood size 2). Learning rates and exchange frequencies are adapted per environment (Table~\ref{tab:hyperparameters_morl}), while the population size is held at 6 across all environments. Reward learning does not terminate after the initial phase: every $K$ steps the pipeline re-queries all teachers, performs 30 online reward-update epochs, and \emph{relabels the shared replay buffer} with the updated reward models, so the policy and the reward models co-adapt throughout training.

\subsection{Policy Set and Evaluation }
\label{app:evalprotocol}
\label{app:policyset}

\paragraph{Weights.} Three weight vectors play distinct roles in LEMUR. The teacher profile $\boldsymbol{\omega}_j \in \Delta^{d}$ is fixed and defines teacher $j$'s utility $u_j = \boldsymbol{\omega}_j^\top \mathbf{r}$. The query weight $\boldsymbol{\omega}^{(n)} \sim \mathrm{Dir}(\kappa\,\boldsymbol{\omega}_j)$ is used only during reward learning. The policy weight $\mathbf{w}_i \in \Delta^{m}$ scalarizes the $m$ learned teacher rewards for policy $\pi_i$; it is assigned by MORL/D, adapted by PSA, and never learned from feedback.

\paragraph{Policy set.} MORL/D maintains $N = 6$ policies $\Pi = \{\pi_i\}_{i=1}^{N}$, each with its own $\mathbf{w}_i$, initialized across $\Delta^{m}$ and adapted by PSA. Following the utility-based view of MORL \cite{roijers2013survey,hayes_practical_2022}, LEMUR outputs the set $\Pi$ rather than a single compromise, and each teacher selects the policy that best serves its own utility.

\paragraph{Return logging.} All reported returns use the native ground-truth reward, which the agent never observes. For each teacher $j$, every training episode of every policy records $G_j(\tau) = \sum_{t} \boldsymbol{\omega}_j^\top \mathbf{r}(s_t,a_t)$. After every 3 episodes, we log $\max G_j(\tau)$ over the most recent 9 episodes, separately per teacher, against environment steps, averaged over five seeds. This is the utility each teacher obtains from the set, and corresponds to the per-profile term of the expected utility metric \cite{felten_multi-objective_2024}.

The two teachers' values may come from different policies. Episodes are stochastic training episodes. Multi-policy methods take the maximum over six policies, single-policy baselines over one \cite{roijers2013survey,teoh2025on}; the comparison with PbMORL, which shares the same six-policy population, is therefore matched. All methods are logged and smoothed identically.
\begin{table*}[tbp]
\centering
\scriptsize
\setlength{\tabcolsep}{4pt}
\renewcommand{\arraystretch}{0.85}
\resizebox{0.92\textwidth}{!}{%
\begin{tabular}{ll|ll}
\toprule
\textbf{Hyperparameter} & \textbf{Value} & \textbf{Hyperparameter} & \textbf{Value} \\
\midrule
\multicolumn{4}{c}{\textit{Stage 1: Unsupervised Pre-training}} \\
Explorer Timesteps & 50,000 (LL), 80,000 (Hopper), 100,000 (Cheetah, MetaWorld) & Explorer Batch Size & 128 \\
Explorer Policy / Q LR & $3 \times 10^{-4}$ & Optimizer & Adam \\
\midrule
\multicolumn{4}{c}{\textit{Stage 2: Reward Learning}} \\
Reward Model & Weight-conditioned MLP & Hidden Layers & 3 $\times$ 256, LeakyReLU($0.2$) \\
Reward Learning Rate & $5\text{e-}4$ (LL), $3\text{e-}4$ (Hopper), $2.5\text{e-}4$ (Cheetah), $1.5\text{e-}4$ (MetaWorld) & Reward Batch Size & 64 \\
Queries per Teacher per Session ($M$) & 200 (LL), 500 (Hopper), 300 (Cheetah, MetaWorld) & Query Length ($H$) & 50 \\
Initial / Online Update Epochs & 100 / 30 & Loss & Bradley-Terry CE \\
Dirichlet Concentration $\kappa$ & 30.0 (All Environments) & Optimizer & Adam \\
Buffer Relabeling & After each reward update & & \\
\bottomrule
\end{tabular}%
}
\caption{LEMUR hyperparameters for unsupervised pre-training (Stage 1) and reward learning (Stage 2). The same query budget, segment length, and reward-learning schedule are used by every preference-based baseline.}
\label{tab:hyperparameters_stages}
\end{table*}

\begin{table*}[tbp]
\centering
\scriptsize
\setlength{\tabcolsep}{4pt}
\renewcommand{\arraystretch}{0.85}
\resizebox{0.92\textwidth}{!}{%
\begin{tabular}{ll|ll}
\toprule
\textbf{Hyperparameter} & \textbf{Value} & \textbf{Hyperparameter} & \textbf{Value} \\
\midrule
Algorithm & MORL/D (MO-SAC) & Batch Size & 512 (LL, MetaWorld), 1024 (Hopper, Cheetah) \\
MORL Timesteps & 100,000 (LL), 750,000 (Hopper, Cheetah), 500,000 (MetaWorld) & Target Entropy Scale & 0.3 \\
Policy LR & $5\text{e-}5$ (LL), $3\text{e-}4$ (Hopper, Cheetah), $1\text{e-}4$ (MetaWorld) & Optimizer & Adam \\
Q LR & $5\text{e-}5$ (LL), $1\text{e-}4$ (Hopper, MetaWorld), $3\text{e-}4$ (Cheetah) & Discount Factor $\gamma$ & 0.99 \\
Population Size & 6 (All Environments) & Shared Buffer & True \\
Exchange Frequency & 5,000 (LL), 10,000 (Hopper, Cheetah, MetaWorld) & Weight Transfer & True \\
Weight Adaptation & PSA & Neighborhood Size & 2 \\
Scalarization & Weighted Sum (ws) & & \\
\bottomrule
\end{tabular}%
}
\caption{LEMUR hyperparameters for multi-objective policy optimization (Stage 3).}
\label{tab:hyperparameters_morl}
\end{table*}

\begin{table*}[tbp]
\centering
\scriptsize
\setlength{\tabcolsep}{4pt}
\renewcommand{\arraystretch}{0.85}
\resizebox{0.92\textwidth}{!}{%
\begin{tabular}{ll|ll}
\toprule
\textbf{Hyperparameter} & \textbf{Value} & \textbf{Hyperparameter} & \textbf{Value} \\
\midrule
\multicolumn{4}{c}{\textit{Shared by all preference-based baselines (Utilitarian, Naive, PbMORL, FPbRL)}} \\
Queries per Teacher per Session ($M$) & As LEMUR (Table~\ref{tab:hyperparameters_stages}) & Query Length ($H$) & 50 \\
Feedback Frequency ($K$) & As LEMUR & Pre-training Buffer & Identical to LEMUR \\
Reward LR / Batch Size / Epochs & As LEMUR & Environment Steps & As LEMUR \\
\midrule
\multicolumn{4}{c}{\textit{Utilitarian and Naive}} \\
Reward Model & Weight-conditioned MLP (as LEMUR) & Policy & Single SAC agent \\
Label Partitioning & Per teacher (Utilitarian); pooled (Naive) & Policy Hyperparameters & Table~\ref{tab:hyperparameters_morl}, population 1 \\
\midrule
\multicolumn{4}{c}{\textit{PbMORL}} \\
Reward Model & Weight-conditioned MLP (as LEMUR) & Population Size & 6 \\
Policy Optimizer & Envelope (LL), MORL/D (others) & & \\
\midrule
\multicolumn{4}{c}{\textit{FPbRL}} \\
Reward Model & GGF welfare MLP ($K$-dim), 256 units & Policy Algorithm & PPO (continuous), SAC-discrete (LL) \\
Policy LR & $3\text{e-}4$ & PPO Steps / Iter & 2,048 \\
PPO Minibatches / Epochs & 32 / 10 & PPO Clip / GAE $\lambda$ & 0.2 / 0.95 \\
GGF Weights & Decreasing on sorted utilities & & \\
\bottomrule
\end{tabular}%
}
\caption{Hyperparameters for the preference-based baselines. All receive the same number of queries per teacher, at the same feedback sessions, as LEMUR.}
\label{tab:hyperparameters_baselines}
\end{table*}

\begin{table*}[tbp]
\centering
\scriptsize
\renewcommand{\arraystretch}{0.9}
\resizebox{0.75\textwidth}{!}{%
\begin{tabular}{ll|ll}
\toprule
\textbf{Hyperparameter} & \textbf{Value} & \textbf{Hyperparameter} & \textbf{Value} \\
\midrule
\multicolumn{4}{c}{\textit{Stage 0 \& 1: Expert Collection \& AIRL Reward Inference}} \\
Expert Timesteps & $1 \times 10^6$ & Demo Episodes & 50 \\
Expert Net Arch. & 256, 256 & Expert Batch Size & 256 \\
Expert Policy / Q LR & $3 \times 10^{-4}$ & Expert Buffer Size & $1 \times 10^6$ \\
Expert $\alpha$ / $\tau$ & 0.2 / 0.005 & Expert Learning Starts & 10,000 \\
AIRL Hidden Dim & 32 & AIRL Generator Timesteps & 200,000 \\
AIRL Disc. Update Interval & 1,024 & AIRL Gen. Net Arch. & 256, 256 \\
AIRL Gen. Buffer / Batch & 300,000 / 256 & Optimizer & Adam \\
\bottomrule
\end{tabular}%
}
\caption{MORAL: hyperparameters for expert demonstration collection and adversarial AIRL reward training.}
\label{tab:moral_stage1}
\end{table*}

\begin{table*}[tbp]
\centering
\scriptsize
\renewcommand{\arraystretch}{0.9}
\resizebox{0.75\textwidth}{!}{%
\begin{tabular}{ll|ll}
\toprule
\textbf{Hyperparameter} & \textbf{Value} & \textbf{Hyperparameter} & \textbf{Value} \\
\midrule
\multicolumn{4}{c}{\textit{Stage 2: Online Policy Optimization with Active Learning}} \\
Policy Algorithm & SAC (all environments; see App.~\ref{app:implementationdetails}) & Active Query Budget & 5,000 \\
Policy / Q LR & $3 \times 10^{-4}$ & Query Interval & 200 \\
Batch Size & 256 & Scalarization Posterior & Bradley-Terry \\
Discount Factor $\gamma$ & 0.99 & Query Strategy & Volume removal \\
\bottomrule
\end{tabular}%
}
\caption{MORAL: hyperparameters for online policy optimization (Stage 2). MORAL's active queries refine a posterior over scalarization weights rather than train a reward model, and are issued in addition to expert demonstrations; they are therefore not directly comparable to LEMUR's per-teacher reward queries. We optimize with SAC rather than the original work's PPO (Appendix~\ref{app:implementationdetails}).}
\label{tab:moral_stage2}
\end{table*}


\section{Baselines Implementation Details}
\label{app:implementationdetails}

To ensure a fair comparison, all baselines share LEMUR's pre-training buffer, scripted teacher profiles, total environment-step budget, and evaluation protocol. Every preference-based baseline (Utilitarian, Naive, PbMORL, and FPbRL) additionally receives exactly the same preference feedback as LEMUR: the same number of queries $M$ per teacher per session, at the same feedback frequency $K$, with the same segment length $H$ (Tables~\ref{tab:hyperparameters_stages} and~\ref{tab:hyperparameters_baselines}). Where a baseline pools feedback, as in Naive and PbMORL-naive, it pools exactly the labels LEMUR receives from all teachers, so no method observes more or fewer preference labels than another. The primary distinction between methods therefore lies in \emph{how conflicting reward signals are aggregated and optimized}, not in the data or compute they receive. The
primary distinction between methods therefore lies in \emph{how conflicting reward signals are
aggregated and optimized}, not in the data or compute they receive.

Each baseline retains its own defining reward-learning mechanism: MORAL's adversarial AIRL
rewards, PbMORL's weight-conditioned Bradley-Terry model, and FPbRL's GGF welfare model. Utilitarian and Naive use the same weight-conditioned reward architecture as LEMUR (Appendix~\ref{app:lemur_implementation}), differing only in how labels are partitioned and how the learned rewards are combined, so that the comparison is fair and not confounded by reward-model capacity. Where
a method's original policy optimizer would place it at an unfair disadvantage in our
environments, we adapt \emph{in the baseline's favour}, upgrading the optimizer rather than
reproducing the paper verbatim, so that reported gaps reflect differences in reward learning
and preference aggregation, the object of study, rather than differences in policy
optimization. Most notably, MORAL is optimized with SAC rather than the PPO used in the
original work, substantially improving its sample efficiency on our continuous-control tasks.
All such deviations are disclosed below.

\paragraph{Utilitarian Agent.}
This baseline imposes an \textit{a priori} scalarization on the objectives. Like LEMUR, it
learns $m$ distinct reward models $\{\hat{r}_{\psi_j}\}_{j=1}^m$, one per teacher, but rather
than learning a set of trade-off policies it trains a \textit{single} SAC agent to optimize
their arithmetic mean:
\begin{equation}
    r_{\text{util}}(s,a) = \frac{1}{m} \sum_{j=1}^m \hat{r}_{\psi_j}(s,a).
\end{equation}
Learning rates, batch sizes, and buffer sizes are identical to LEMUR, with the population size
set to $1$ (standard single-objective SAC).

\paragraph{Naive Data Pooling.}
This baseline aggregates conflicting preferences at the \emph{data} level, mimicking standard
RLHF applied to heterogeneous feedback. Rather than maintaining separate datasets
$\mathcal{D}_j$ per objective, all feedback tuples are stored in a single monolithic dataset
$\mathcal{D}_{\text{pool}} = \bigcup_{j=1}^m \mathcal{D}_j$, over which one reward model
$\hat{r}_{\text{pool}}$ is trained to minimize the cross-entropy loss. The policy is then
trained with standard SAC to maximize
\begin{equation}
    r_{\text{naive}}(s,a) = \hat{r}_{\text{pool}}(s,a).
\end{equation}
As with the Utilitarian baseline, we use the environment-specific hyperparameters in
Table~\ref{tab:hyperparameters_morl} with the population size fixed to $1$.

\paragraph{PbMORL}
The original Pb-MORL assumes a single, internally-consistent teacher whose preferences are valid
under any sampled scalarization weight, and therefore has no mechanism for multiple,
independently-opinionated teachers. We retain its weight-conditioned $m$-dimensional reward
model, trained with the Bradley-Terry cross-entropy objective over pairwise trajectory
comparisons, and extend it to our setting in the two most natural ways: \emph{PbMORL-naive}
pools all teachers' preferences into one shared weight-conditioned model, while
\emph{PbMORL-utilitarian} trains a separate model per teacher and combines them at inference by
a fixed uniform average. Each teacher answers every query under its own fixed weight vector,
exactly as in LEMUR.

The original paper pairs its reward model with Envelope Q-learning,
a discrete, value-based method. We retain this paper-faithful pairing on LunarLander, but
Envelope's Q-network requires an argmax over actions and is structurally inapplicable to
continuous control; on Hopper, HalfCheetah, and MetaWorld we therefore substitute
MORL/D, the same population-based optimizer LEMUR uses. This is the only
structural substitution, and it equalizes the optimizer between PbMORL and LEMUR on those
environments, isolating the reward-learning strategy as the sole difference.

\paragraph{FPbRL}
We implement FPbRL's $K$-dimensional welfare reward model, trained from Generalized Gini
Welfare (GGF)-based preferences over the same $K=2$ scripted teachers across which LEMUR and the
other baselines compromise. At each policy-update iteration, the learned $K$-dimensional reward
is scalarized by the GGF weight assignment, which places the largest weight on the currently
worst-off objective, refreshed from a running per-objective return estimate, the standard
practical form of Siddique et al.'s fair policy gradient.

Unlike MORAL, we retain PPO for continuous-action environments, matching the original paper:
FPbRL's fair policy gradient is formulated for an on-policy optimizer, and substituting an
off-policy method would alter the method's semantics rather than simply strengthen it. On
LunarLander we use SAC-discrete, as no discrete PPO implementation is used elsewhere in our
pipeline. In addition to the per-teacher returns reported for all methods, we log FPbRL's own
fairness metrics (welfare, coefficient of variation, and minimum objective), so that it is also
evaluated on the criterion it is explicitly designed to optimize.

\paragraph{MORAL}
We reimplement MORAL's pipeline in three stages: (i) two expert policies are trained on the
ground-truth per-teacher scalarized reward and used to collect demonstrations; (ii) for each
teacher, an AIRL discriminator is trained \emph{adversarially} against a
live generator, rather than against a fixed pool of shuffled expert transitions, which would
render training non-adversarial, yielding a per-teacher learned reward $g(s)$; and (iii) an
active-learning wrapper scalarizes the two frozen AIRL rewards with a Bradley-Terry weight
posterior, updated online via volume-removal preference queries, which a single-objective
policy optimizes. Since the original MORAL was evaluated only on grid-world tasks, we adopt the
AIRL hyperparameters and network architectures of Fu et al. for our
high-dimensional MuJoCo experiments; these are detailed in Tables~\ref{tab:moral_stage1}
and~\ref{tab:moral_stage2}.

We deliberately preserve MORAL's defining contribution, adversarial IRL reward learning
combined with an actively-queried scalarization posterior, and do not replace it with our own
preference-based reward model, as doing so would no longer constitute a MORAL baseline. We do,
however, strengthen its policy optimization: the original work uses PPO, whereas we optimize
with SAC (SAC-continuous, or SAC-discrete on LunarLander) across all environments. This
off-policy upgrade materially improves MORAL's sample efficiency under an identical
environment-step budget and matches the backbone family used by LEMUR, ensuring MORAL is not
penalized for an on-policy optimizer choice unrelated to its reward-learning contribution.

\subsection{MORAL Baseline: Learned Weights and Optimiser Choice}
\label{app:moralsacvsppo}

\paragraph{Learned scalarisation weights.}
MORAL maintains a posterior over a scalarisation weight vector, refined through active queries,
and it is this posterior, rather than a per-teacher reward decomposition, that carries its
notion of whose preferences the policy is serving. Figure~\ref{fig:moral_weights} tracks both
components over training.
\paragraph{SAC vs.\ PPO.}
As described in Appendix~\ref{app:implementationdetails}, we optimize the MORAL baseline with SAC
rather than the PPO used in the original work, on the grounds that an on-policy optimizer would
disadvantage the baseline for reasons unrelated to its reward-learning contribution. This
ablation verifies that the substitution is genuinely favourable to MORAL and therefore that our
reported comparison is conservative.

\begin{figure}[tbp]
    \centering
    \includegraphics[width=0.65\linewidth]{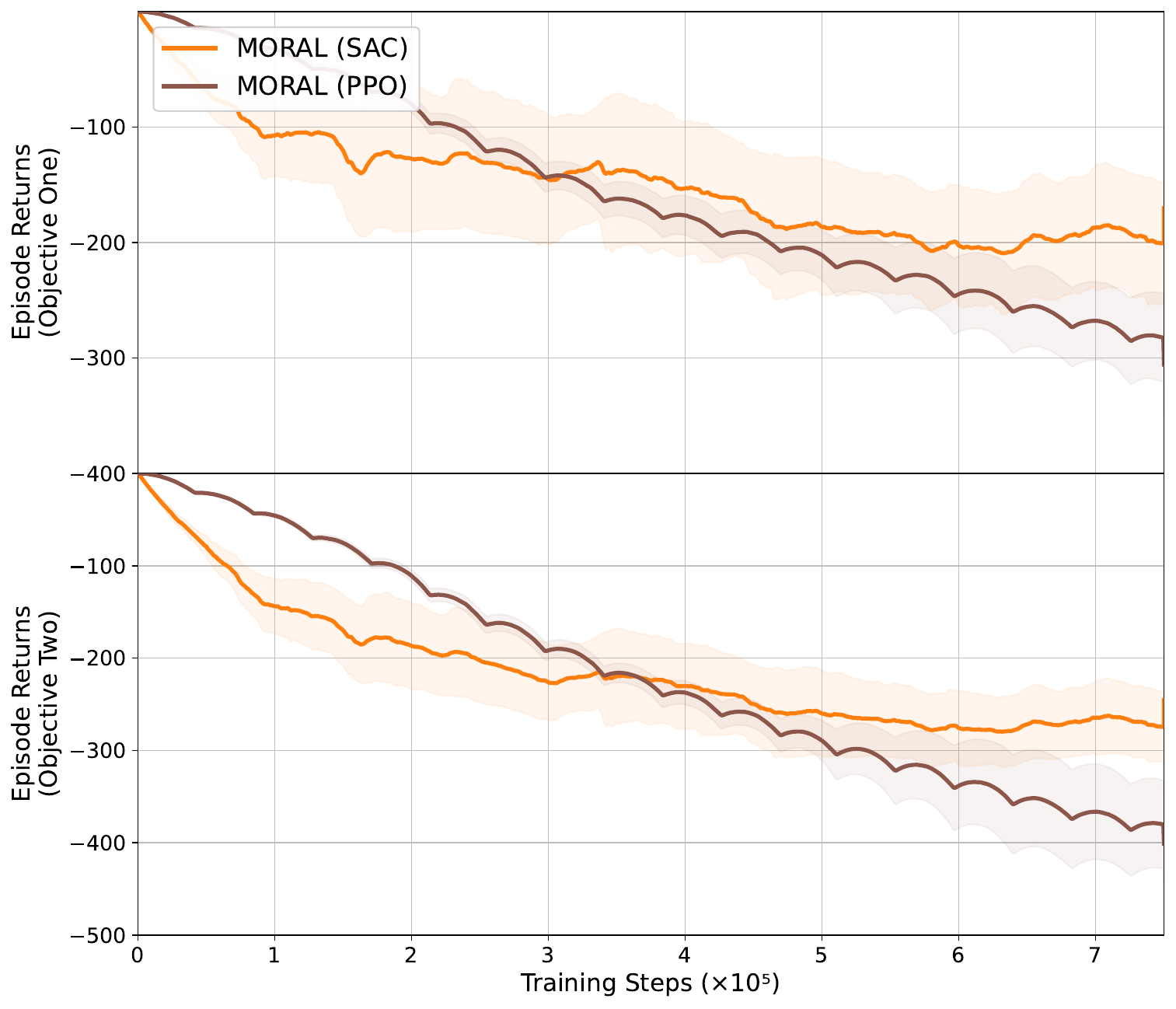}
    \caption{\textbf{MORAL optimizer ablation (\texttt{MO-Cheetah}).} With MORAL's reward
    learning held fixed, SAC outperforms the original work's PPO on both objectives, confirming
    that our SAC upgrade strengthens the baseline. Curves are heavily smoothed to expose the
    trend beneath PPO's on-policy oscillation. Mean $\pm$ std over five seeds.}
    \label{fig:moral_sac_vs_ppo}
\end{figure}

Figure~\ref{fig:moral_sac_vs_ppo} confirms this: holding MORAL's adversarial AIRL reward learning
and active-query scalarisation posterior fixed and varying only the policy optimizer, the SAC
variant dominates PPO on both objectives throughout training, and the gap widens as training
proceeds. The PPO variant additionally displays a pronounced sawtooth characteristic of
on-policy updates (shown here under heavy smoothing). Reporting MORAL with SAC therefore
strengthens the baseline relative to a faithful reproduction of the original paper, and the
weight collapse in Figure~\ref{fig:moral_weights} is a property of MORAL's scalarization
posterior rather than a symptom of the optimizer.

\begin{figure}[tbp]
    \centering
    \includegraphics[width=0.85\linewidth]{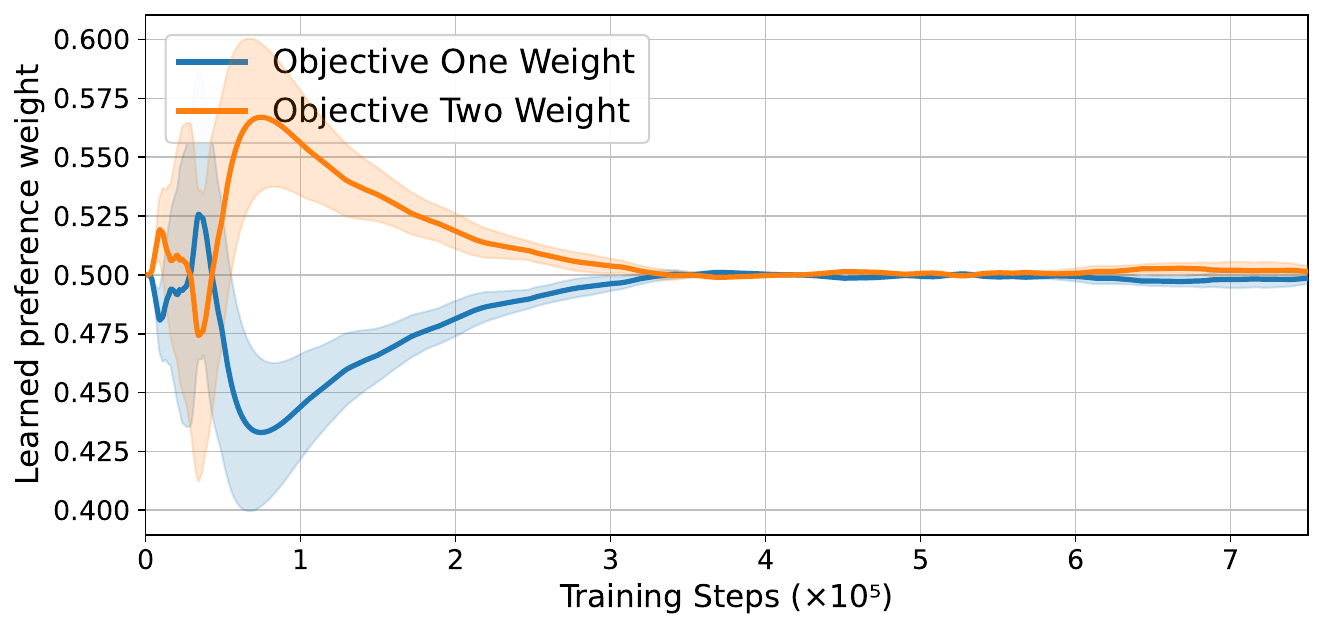}
    \caption{\textbf{MORAL's learned scalarisation weights in
    (\texttt{MO-Cheetah}).} The two components of MORAL's active-query scalarisation posterior
    over training.}
    \label{fig:moral_weights}
\end{figure}

\section{Additional Results}
\subsection{Multi-Objective RL Metrics}
\label{app:additionalresults}

Evaluating a multi-objective agent requires assessing the \emph{set} of policies it recovers rather than any single return, so we adopt two metrics standard in the MORL literature. Let $\mathcal{P} = \{J(\pi_1), \dots, J(\pi_P)\}$ denote the set of objective-value vectors attained by the learned policy population.

\paragraph{Hypervolume (HV).} Given a reference point $\mathbf{z}$ dominated by all solutions, Hypervolume is the volume of the region dominated by $\mathcal{P}$ and bounded by $\mathbf{z}$,
\begin{equation}
    \mathrm{HV}(\mathcal{P}, \mathbf{z}) = \Lambda\!\left( \bigcup_{\mathbf{p} \in \mathcal{P}} [\mathbf{z}, \mathbf{p}] \right),
\end{equation}
where $\Lambda$ is the Lebesgue measure. It simultaneously rewards solutions that are high-performing (pushing the front outward) and diverse (covering more of the objective space), and is the most widely adopted MORL quality indicator because it is the only common metric strictly monotonic with Pareto dominance: any set that dominates another is guaranteed a higher score. We use the reference point supplied by the \textit{MORL-Generalization} benchmark so that values are comparable across methods within an environment; absolute magnitudes are not comparable \emph{across} environments, since they depend on both the reference point and the reward scale.

\paragraph{Sparsity (SPS).} Sparsity measures how evenly solutions are distributed along the recovered front. Sorting the $|\mathcal{P}|$ solutions by each objective $i$ and writing $\tilde{P}_i(k)$ for the $k$-th value,
\begin{equation}
    \mathrm{SPS}(\mathcal{P}) = \frac{1}{|\mathcal{P}| - 1} \sum_{i=1}^{m} \sum_{k=1}^{|\mathcal{P}|-1} \left( \tilde{P}_i(k) - \tilde{P}_i(k+1) \right)^2,
\end{equation}
with lower values indicating more uniform coverage. Sparsity must be read alongside Hypervolume rather than independently: a degenerate front that collapses to a single solution reports a low, apparently favourable value despite failing to cover the objective space, which is why we report both. This is the case for FPbRL, whose fixed welfare scalarization converges to a single policy, so no front is recovered and sparsity is undefined ($-$) on three of four environments.

\begin{table}[htbp]
  \centering
  \small
  \setlength{\tabcolsep}{4pt}
  \resizebox{\columnwidth}{!}{%
    \begin{tabular}{lccc}
      \toprule
      \textbf{Environment} & \textbf{LEMUR} & \textbf{PbMORL} & \textbf{FPbRL} \\
      \midrule
      \multicolumn{4}{c}{\textbf{Hypervolume ($\uparrow$)}} \\
      \midrule
      LunarLander           & $\mathbf{1.10\times10^4 \pm 3.48\times10^2}$ & $1.09\times10^4 \pm 5.74\times10^2$          & $5.05\times10^3 \pm 1.28\times10^3$ \\
      Hopper                & $\mathbf{3.67\times10^6 \pm 2.84\times10^5}$ & $2.27\times10^6 \pm 3.11\times10^4$          & $7.19\times10^4 \pm 4.57\times10^4$ \\
      HalfCheetah           & $4.86\times10^7 \pm 1.34\times10^6$          & $\mathbf{4.90\times10^7 \pm 1.18\times10^6}$ & $7.73\times10^5 \pm 4.12\times10^5$ \\
      MetaWorld-DrawerClose & $2.40\times10^6 \pm 9.71\times10^5$          & $\mathbf{3.03\times10^6 \pm 2.25\times10^6}$ & $2.90\times10^6 \pm 2.05\times10^6$ \\
      \midrule
      \multicolumn{4}{c}{\textbf{Sparsity ($\downarrow$)}} \\
      \midrule
      LunarLander           & $\mathbf{3.1 \pm 2.7}$       & $4.1 \pm 1.4$      & $-$ \\
      Hopper                & $\mathbf{250.1 \pm 5.8}$     & $1067.6 \pm 355.3$ & $-$ \\
      HalfCheetah           & $\mathbf{3001.8 \pm 1580.6}$ & $4128.7 \pm 2825.5$ & $-$ \\
      MetaWorld-DrawerClose & $\mathbf{428.6 \pm 422.9}$   & $3814.4 \pm 3619.3$ & $-$ \\
      \bottomrule
    \end{tabular}%
  }
  \caption{Hypervolume and Sparsity (mean $\pm$ standard deviation over seeds) across benchmark environments. FPbRL optimizes a single fixed welfare-scalarized policy and therefore returns one solution rather than a front, so Sparsity is undefined ($-$).}
  \label{tab:morl_metrics_full_single_col}
\end{table}

\subsection{Reward Model Evaluation Metrics}
\label{app:reward_metrics}

Policy return alone cannot distinguish a reward model that has genuinely recovered a teacher's utility from one merely correlated with it on the visited state distribution. We therefore evaluate the learned reward models directly against the ground-truth teacher utilities, following PbRL benchmarking practice. Let $\hat{r}_{\psi_j}(s,a) = \boldsymbol{\omega}_j^\top \hat{\mathbf{R}}_{\psi_j}(s,a)$ denote teacher $j$'s learned scalar reward and $u_j(s,a) = \boldsymbol{\omega}_j^\top \mathbf{r}(s,a)$ its ground-truth utility, where $\mathbf{r}$ is the environment's native $d$-dimensional reward and $\boldsymbol{\omega}_j$ teacher $j$'s scripted profile. All metrics lie in $[-1,1]$, higher is better, and are averaged across the $m$ teachers.

\paragraph{Per-state correlation.} Over states sampled from evaluation rollouts, \emph{Spearman} rank correlation measures how faithfully the learned reward orders individual transitions. It is our primary metric because preference-based rewards are identifiable only up to a positive monotone transform, making an order-preserving measure the appropriate notion of correctness. We also report \emph{Pearson} correlation on raw values, which is stricter in penalising any nonlinear distortion; following prior reward-evaluation work we refer to this as the \emph{Value-Order Correlation} (VOC).

\paragraph{Trajectory and policy ranking.} Per-state correlation can be high even when a reward model induces the wrong ordering over whole \emph{behaviours}, which is what the policy ultimately optimizes. We therefore roll out each policy in the MORL/D population and compare the induced rankings using Kendall's $\tau$-b over trajectory returns, and the \emph{Trajectory Alignment Coefficient} (TAC), computed identically but on \emph{discounted} returns so that alignment is weighted by the same temporal discounting the agent optimizes. \emph{Trajectory VOC} additionally averages the within-trajectory correlation between learned and ground-truth per-step reward sequences, capturing whether the model tracks the shape of the signal within an episode rather than only across episodes.

\begin{table}[htbp]
  \centering
  \resizebox{\columnwidth}{!}{%
    \begin{tabular}{lccc}
      \toprule
      \textbf{Metric} & \textbf{LEMUR} & \textbf{PbMORL} & \textbf{FPbRL} \\
      \midrule
      Spearman ($\rho \uparrow$)           & $\mathbf{0.710 \pm 0.005}$ & $0.701 \pm 0.004$          & $0.512 \pm 0.089$ \\
      Pearson ($r \uparrow$)               & $0.863 \pm 0.003$          & $\mathbf{0.870 \pm 0.017}$ & $0.493 \pm 0.065$ \\
      VOC ($\uparrow$)                     & $0.863 \pm 0.003$          & $\mathbf{0.870 \pm 0.017}$ & $0.493 \pm 0.065$ \\
      VOC (traj) ($\uparrow$)              & $0.858 \pm 0.003$          & $\mathbf{0.866 \pm 0.016}$ & $0.494 \pm 0.066$ \\
      Kendall $\tau$ (policy) ($\uparrow$) & $0.939 \pm 0.030$          & $\mathbf{0.945 \pm 0.012}$ & $0.567 \pm 0.233$ \\
      Traj-Alignment Coeff ($\uparrow$)    & $0.898 \pm 0.041$          & $\mathbf{0.903 \pm 0.045}$ & $0.633 \pm 0.233$ \\
      \bottomrule
    \end{tabular}%
  }
  \caption{Reward-model evaluation metrics: MO-Cheetah. Mean $\pm$ standard deviation over seeds.}
  \label{tab:reward_eval_halfcheetah}
\end{table}
\begin{table}[htbp]
  \centering
  \resizebox{\columnwidth}{!}{%
    \begin{tabular}{lccc}
      \toprule
      \textbf{Metric} & \textbf{LEMUR} & \textbf{PbMORL} & \textbf{FPbRL} \\
      \midrule
      Spearman ($\rho \uparrow$)         & $\mathbf{0.520 \pm 0.007}$ & $0.231 \pm 0.000$ & $0.085 \pm 0.074$ \\
      Pearson ($r \uparrow$)            & $\mathbf{0.589 \pm 0.020}$ & $0.335 \pm 0.000$ & $0.345 \pm 0.005$ \\
      VOC ($\uparrow$)                  & $\mathbf{0.589 \pm 0.020}$ & $0.335 \pm 0.000$ & $0.345 \pm 0.005$ \\
      VOC (traj) ($\uparrow$)           & $0.340 \pm 0.035$          & $0.339 \pm 0.000$ & $\mathbf{0.358 \pm 0.010}$ \\
      Kendall $\tau$ (policy) ($\uparrow$) & $\mathbf{0.359 \pm 0.002}$ & $0.291 \pm 0.000$ & $-0.167 \pm 0.100$ \\
      Traj-Alignment Coeff ($\uparrow$) & $\mathbf{0.347 \pm 0.026}$ & $0.300 \pm 0.000$ & $0.167 \pm 0.167$ \\
      \bottomrule
    \end{tabular}%
  }
  \caption{Reward-Model Evaluation Metrics: MO-MetaWorld. Mean $\pm$ standard deviation over seeds.}
  \label{tab:reward_eval_drawerclose}
\end{table}

\begin{table}[htbp]
  \centering
  \resizebox{\columnwidth}{!}{%
    \begin{tabular}{lccc}
      \toprule
      \textbf{Metric} & \textbf{LEMUR} & \textbf{PbMORL} & \textbf{FPbRL} \\
      \midrule
      Spearman ($\rho \uparrow$)           & $\mathbf{0.945 \pm 0.002}$ & $0.512 \pm 0.015$ & $0.648 \pm 0.094$ \\
      Pearson ($r \uparrow$)               & $\mathbf{0.951 \pm 0.002}$ & $0.518 \pm 0.014$ & $0.716 \pm 0.073$ \\
      VOC ($\uparrow$)                     & $\mathbf{0.951 \pm 0.002}$ & $0.518 \pm 0.014$ & $0.716 \pm 0.073$ \\
      VOC (traj) ($\uparrow$)              & $\mathbf{0.947 \pm 0.004}$ & $0.585 \pm 0.020$ & $0.714 \pm 0.073$ \\
      Kendall $\tau$ (policy) ($\uparrow$) & $0.914 \pm 0.002$          & $0.347 \pm 0.057$ & $\mathbf{0.933 \pm 0.000}$ \\
      Traj-Alignment Coeff ($\uparrow$)    & $0.856 \pm 0.014$          & $0.343 \pm 0.026$ & $\mathbf{0.933 \pm 0.000}$ \\
      \bottomrule
    \end{tabular}%
  }
  \caption{Reward-model evaluation metrics: Hopper. Mean $\pm$ standard deviation over seeds.}
  \label{tab:reward_eval_hopper}
\end{table}

\begin{table*}[tbp]
\centering
\scriptsize
\setlength{\tabcolsep}{4pt}
\renewcommand{\arraystretch}{1.0}
\resizebox{\textwidth}{!}{%
\begin{tabular}{lllll}
\toprule
\textbf{Environment} & \textbf{Actions} & \textbf{$d$} & \textbf{Native Components} & \textbf{Teacher Profiles $(\boldsymbol{\omega}_A;\ \boldsymbol{\omega}_B)$} \\
\midrule
\texttt{MO-LunarLander} & Discrete & 4 & shaping, main-engine cost, side-engine cost, landing & $[0.6, 0.3, 0.05, 0.05]$;\ \ $[0.6, 0.05, 0.3, 0.05]$ \\
\texttt{MO-Hopper} & Continuous & 3 & forward velocity, jump height, energy cost & $[0.8, 0.1, 0.1]$;\ \ $[0.3, 0.5, 0.2]$ \\
\texttt{MO-Cheetah} & Continuous & 2 & forward velocity, control cost & $[0.6, 0.4]$;\ \ $[0.4, 0.6]$ \\
\texttt{MO-MetaWorld (Drawer-Close)} & Continuous & 2 & task progress, control effort & $[0.6, 0.4]$;\ \ $[0.4, 0.6]$ \\
\bottomrule
\end{tabular}%
}
\caption{Benchmark environments: native reward decomposition and scripted teacher profiles. $d$ denotes the dimensionality of the native vector reward; all environments use $m=2$ teachers.}
\label{tab:env_details}
\end{table*}
\paragraph{Results.} Tables~\ref{tab:reward_eval_halfcheetah}-\ref{tab:reward_eval_hopper} report all metrics. Tables~\ref{tab:reward_eval_halfcheetah}--\ref{tab:reward_eval_hopper} report all metrics. LEMUR attains substantially stronger correlations than both baselines on Hopper and MetaWorld, where the teachers' profiles conflict most, and performs on par with PbMORL on Cheetah, where the two methods' returns are also closest. Trajectory-level metrics are more mixed: FPbRL attains a higher Kendall $\tau$ and TAC on Hopper despite substantially weaker per-state correlation, reflecting that its welfare scalarization orders whole behaviours consistently even where the underlying reward is poorly calibrated. Reporting both families is what makes this distinction visible, and we recommend the same practice for future work in this setting.


\section{Benchmark Environment Details}
\label{app:envdetail}

Table~\ref{tab:env_details} summarises the native multi-objective structure of each benchmark
environment and the scripted teacher profiles used throughout our experiments. In every
case, teacher $j$'s ground-truth utility is the linear scalarisation
$u_j(s,a) = \boldsymbol{\omega}_j^\top \mathbf{r}(s,a)$ of the environment's native $d$-dimensional
reward vector $\mathbf{r}$, and it is this quantity that
\texttt{validation/gt\_teacher\_\{a,b\}\_return} tracks during training.

The profiles define the scripted teachers: each teacher's utility is a linear scalarization of the environment's native reward components, the standard utility model in MORL \cite{roijers2013survey,hayes_practical_2022}, and labels are generated from ground-truth rewards following the standard PbRL benchmarking protocol \cite{lee2021bpref}. The values are set by us to induce conflict between teachers, and every baseline receives the same profiles.

The profiles are chosen
to be genuinely conflicting: each teacher places its largest weight on a different component, so
no single policy can simultaneously maximise both utilities.

Our conclusions do not depend on these particular values: Appendix~\ref{app:conflict_levels} varies the conflict on \texttt{MO-Cheetah} from $[0.6, 0.4]$ to $[0.9, 0.1]$, and Appendix~\ref{app:overlapping} considers near-agreement.

\paragraph{MO-LunarLander.} A four-component variant of the classic LunarLander domain with a
discrete action space, taken from the MORL-Generalization benchmark \cite{teoh2025on}. The
native reward vector comprises the shaping term (progress toward the landing pad), the main-engine
fuel cost, the side-engine fuel cost, and the terminal landing/crash outcome. Teacher A weights
the two engine-cost terms asymmetrically against Teacher B, producing a fuel-allocation conflict
on top of a shared landing objective.

\paragraph{MO-Hopper.} A three-component continuous-control locomotion task in which the native
reward vector is $[\,v_x,\ h,\ -c\,\|a\|^2\,]$: forward velocity, hop height, and a negated
energy/control cost. Teacher A ($\boldsymbol{\omega}_A = [0.8, 0.1, 0.1]$) strongly prefers fast
locomotion, while Teacher B ($\boldsymbol{\omega}_B = [0.3, 0.5, 0.2]$) prefers a higher, more
energy-efficient and stable gait.

\paragraph{MO-Cheetah.} A two-component continuous-control task whose native reward vector is
$[\,\texttt{reward\_forward},\ \texttt{reward\_ctrl}\,]$, i.e.\ forward velocity against control
cost. The teacher profiles $[0.6, 0.4]$ and $[0.4, 0.6]$ place opposing emphasis on velocity
versus energy efficiency.

\paragraph{MO-MetaWorld (Drawer-Close).} Meta-World tasks are natively \emph{single}-objective,
providing only a dense task-progress reward. We convert \texttt{Drawer-Close} into a
two-component task by pairing this native reward with a control-effort penalty, yielding the
vector reward $[\,r_{\text{task}},\ -\lambda\|a\|_2^2\,]$ with control-cost weight
$\lambda = 0.1$ and a fixed horizon of 500 steps. This mirrors the forward-reward/control-cost
decomposition standard to the MuJoCo suite, but applied to a more complex robot-manipulation
domain. Teacher A is rewarded by task completion, Teacher B by smooth, energy-efficient
actuation.
\paragraph{Scripted teacher labelling.}
\label{app:scriptedteachers}
Each teacher $j$ is defined by its profile $\boldsymbol{\omega}_j$ (Table~\ref{tab:env_details}). Writing $\mathbf{R}(\sigma) = \sum_t \mathbf{r}(s_t, a_t)$ for the native vector return of a segment, teacher $j$ answers the $n$-th query under its sampled weight $\boldsymbol{\omega}^{(n)} \sim \mathrm{Dir}(\kappa\,\boldsymbol{\omega}_j)$ with
\begin{equation}
\label{eq:scriptedteacher}
y^{(n)} =
\begin{cases}
1, & \boldsymbol{\omega}^{(n)\top}\mathbf{R}(\sigma^1) > \boldsymbol{\omega}^{(n)\top}\mathbf{R}(\sigma^0),\\[2pt]
0, & \boldsymbol{\omega}^{(n)\top}\mathbf{R}(\sigma^1) < \boldsymbol{\omega}^{(n)\top}\mathbf{R}(\sigma^0),\\[2pt]
0.5, & \text{otherwise.}
\end{cases}
\end{equation}
The agent observes only the tuple $(\sigma^0, \sigma^1, \boldsymbol{\omega}^{(n)}, y^{(n)})$. The native reward $\mathbf{r}$ and the teacher utilities $u_j$ (Equation~\ref{eq:teacherutility}) are used solely to generate labels and to evaluate the learned models, and are never available to any method. To verify the robustness of LEMUR to noise, in the ablation experiments in Figure \ref{fig:three_ablations}, each label is independently flipped with a fixed probability before being returned to the agent, following PbRL benchmarking protocol \cite{lee2021bpref}.

\section{Human Experiment Protocol}
\label{app:humanprotocol}
Teachers labeled pairs of 50-step trajectory segments, shown side by side as looping videos in a browser interface, by selecting the preferred segment, marking both as equally preferable, or skipping incomparable pairs \cite{lee2021bpref}. Following LEMUR's interactive loop (Algorithm~\ref{alg:LEMUR}), feedback was collected over 8 sessions of 25 queries per teacher (200 queries each), with Teacher A answering all queries of a session before Teacher B, followed by reward learning and 60k steps of policy optimization. Queries were sampled uniformly from the pre-training rollouts and shared by both teachers.

\section{Compute Resources}
In all experiments, we use 12 CPUs and a single GPU, of type either NVIDIA A100 or L40. Training in all environments takes approximately three to five hours on average.

\end{document}